\documentclass[lettersize,journal]{IEEEtran}
\usepackage{amsmath,amsfonts}
\usepackage{algorithm}
\usepackage{array}
\usepackage[caption=false,font=normalsize,labelfont=sf,textfont=sf]{subfig}
\usepackage{textcomp}
\usepackage{stfloats}
\usepackage{url}
\usepackage{verbatim}
\usepackage{graphicx}
\usepackage{cite}
\usepackage{picinpar}
\usepackage{flushend}
\usepackage[utf8]{inputenc}
\usepackage{colortbl}
\usepackage{soul}
\usepackage{multirow}
\usepackage{pifont}
\usepackage{color}
\usepackage{alltt}
\usepackage[hidelinks]{hyperref}
\usepackage{enumerate}
\usepackage{siunitx}
\usepackage{breakurl}
\usepackage{epstopdf}
\usepackage{pbox}
\usepackage{amssymb}
\usepackage{booktabs}
\usepackage{balance}
\usepackage{makecell}
\usepackage{amsthm} 
\usepackage{amsmath,amssymb,amsfonts,amsthm}
\usepackage{bm}
\usepackage{algpseudocode}
 \usepackage{tabularx}
\usepackage{amsmath,amssymb}
\usepackage{algorithm}       
\usepackage{algpseudocode}

\newtheorem{tm}{Theorem}

\begin{document}

\title{NSFlow: End-to-End Differentiable Neuro-Symbolic Optical Flow for Visual Odometry}

\author
{
	\vskip 1em
	Yicheng Lin$^{2, *}$,
    Zhipeng Fei$^{1, *}$,
    Yuxiu Xu$^{1}$,
    WenDong Chen$^{1}$,
    Cong Li$^{1}$,
	and Bin Han$^{1}$, \emph{Senior Member, IEEE}

	\thanks{

        This work was supported National Nature Science Foundation of China (52375015) and in part by the Interdisciplinary Research Program of HUST (2024JCYJ037).  (Yicheng Lin and Zhipeng Fei contributed equally to this work.)(Corresponding author: Bin Han.) \par
    
        $^{1}$Z. Fei, H. Xie, C. Li, and B. Han are with the State Key Laboratory of Intelligent Manufacturing Equipment and Technology, School of Mechanical Science and Engineering, Huazhong University of Science and Technology, Wuhan, China. (email:{\tt\footnotesize binhan@hust.edu.cn})
    
        $^{2}$Y. Lin is with Guangzhou Purpleriver Electronic Technology Co., Ltd, Guangzhou, China.%

            }
}

\maketitle

\begin{abstract}

Sparse optical flow provides stable inter-frame correspondence, playing a key role in Visual Odometry (VO) and Visual-Inertial Odometry (VIO). Classical optimization-based methods, such as Lucas-Kanade (LK), perform well under small displacements but are sensitive to large motions and illumination changes. Modern regression-based learning methods, while more robust in complex scenes, are often computationally heavy and lack explicit geometric consistency, making them less suitable for efficient VO/VIO front-ends. To bridge this gap, we propose a hybrid neuro-symbolic framework that combines the strengths of both paradigms. Our method uses a Convolutional Neural Network (CNN) to extract robust feature representations, which is fed into a differentiable LK optimizer to estimate optical flow in an end-to-end trainable manner. Through implicit differentiation, gradients are propagated across the iterative solver, enabling joint optimization of feature extraction and flow estimation. The resulting system integrates seamlessly into existing VO/VIO pipelines and runs in real-time on embedded platforms. Experiments show that our method outperforms conventional optimization-based flow in challenging conditions such as dynamic lighting and low texture, while also achieving higher accuracy and lower latency than purely regression-based alternatives. When deployed in a VIO system, our method demonstrates significant performance improvement, achieving an average error reduction of 42\% on challenging datasets while enhancing tracking stability. The code is publicly available.

\end{abstract}


\section{Introduction}


\IEEEPARstart{V}{isual} Odometry (VO) and Visual-Inertial Odometry (VIO) are techniques that estimate a device's or robot's own movement trajectory in environments where GPS is unavailable. This forms a core capability for autonomous navigation \cite{scaramuzza2011visual,forster2017svo}. The accuracy and stability of these systems depend heavily on inter-frame correspondence—the process of matching pixels or features between consecutive images to provide geometric constraints for motion estimation \cite{mur2015orb,qin2018vins}. However, in real-world conditions, factors such as, changing lighting, motion blur, and low-texture areas can severely degrade the quality of these matches. This leads to accumulated drift in the estimated trajectory and, ultimately, a loss of localization stability \cite{engel2018direct,chen2023vist}. Therefore, achieving robust and reliable inter-frame correspondence under challenging visual conditions remains a key obstacle to the practical deployment of VO/VIO systems.\par 


Image correspondence methods are generally categorized into direct and indirect approaches \cite{engel2014lsd,engel2018direct}. Direct methods perform image registration by minimizing photometric errors at the pixel level. This approach typically relies on the assumptions of brightness constancy and the presence of sufficient texture. Indirect methods follow a two-step process: establishing feature correspondences or optical flow first, then leveraging them for geometric optimization. They are typically regarded as more adaptable to diverse and difficult environments as a result \cite{mur2015orb,forster2017svo}. Here, we focus on optical flow, which is based on the principle of video continuity to measure the subtle, sub-pixel displacements of keypoints across frames. This enables robust and continuous pose estimation. Hence, optical flow is a popular choice for correspondence in systems ranging from classics like VINS-Mono \cite{qin2018vins} and MSCKF \cite{mourikis2007msckf} to modern learning-based models like DROID\cite{teed2021droid,tang2019ba}. Yet, despite its widespread use, its robustness can be severely compromised by illumination variations, large displacements, and the constraints of real-time computation. \par 

\begin{figure}[!t]
        \centering
        \includegraphics[width=0.48\textwidth]{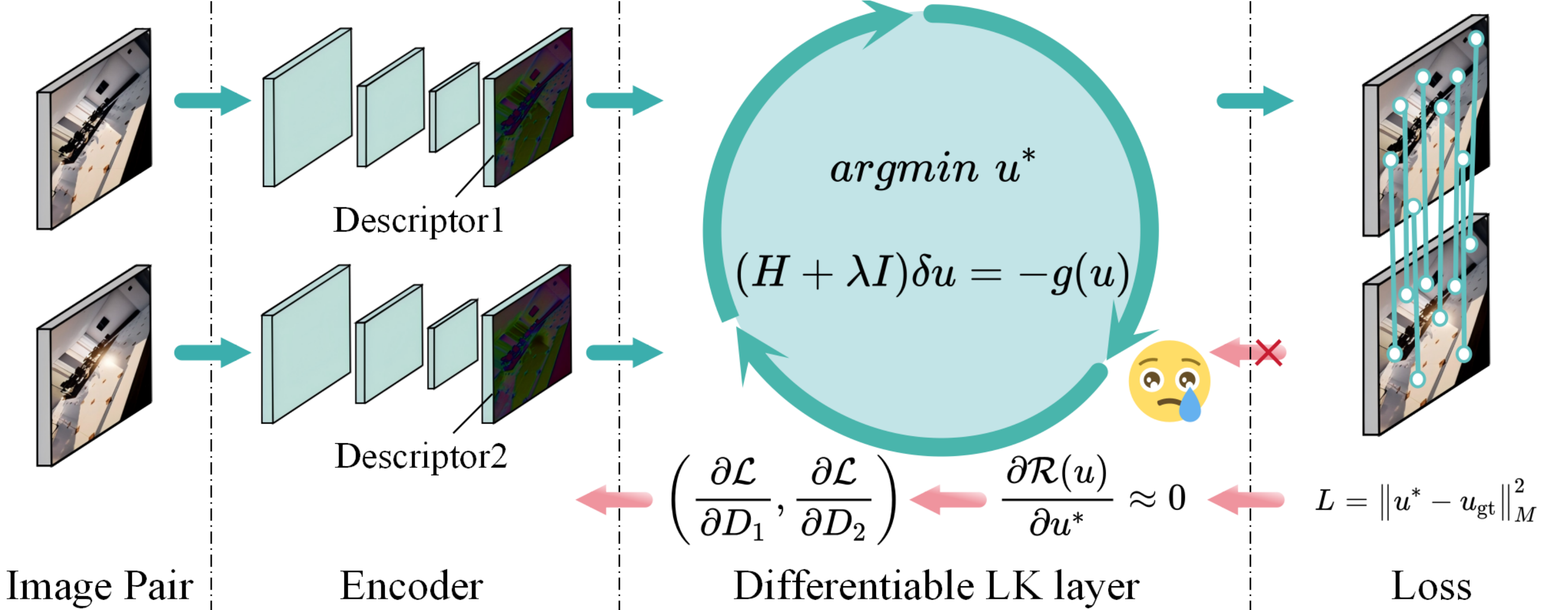}
        \caption{\textbf{A Framework for End-to-End Trained Neuro-Symbolic Optical Flow.} In the forward pass, a convolutional network extracts image features for an iteratively optimized LK algorithm to solve optical flow; during backpropagation, gradients through this optimization are computed via the implicit function theorem to enable error propagation and joint training.} 
        \label{for-backward} 
        \end{figure}

Optical flow estimation methods can be broadly categorized into optimization-based and regression-based paradigms. Optimization-based methods (e.g., Lucas-Kanade) estimate motion by solving an energy function derived from photometric and spatial constraints \cite{horn1981optical,lk_flow}. This formulation ensures efficiency and geometric consistency, but its reliance on explicit assumptions (e.g., brightness constancy) limits robustness under large motions or illumination changes. In contrast, regression-based methods use deep networks to directly predict flow from image pairs. Trained on large datasets, they capture complex motion patterns and show strong robustness to appearance variations \cite{flownet,flownet2,raft}. However, as noted in \cite{teed2021droid}, their outputs often lack explicit geometric consistency, making them less suitable for visual odometry without post-processing. Recent hybrid approaches integrate global optimization (e.g., bundle adjustment) into learning frameworks to enhance geometric consistency \cite{teed2021droid,dpvo,tartanvo}. Yet, these iterative refinements incur high computational and memory costs, leaving a critical trade-off among robustness, accuracy, and efficiency—especially for real-time embedded VO/VIO systems. \par 

To address this trade-off, we introduce a neuro-symbolic learning framework for end-to-end trainable sparse optical flow, as shown in Fig.\ref{for-backward}. As introduced in \cite{neuro}, neuro-symbolic learning refers to a category of methods that embed interpretable symbolic modules into data-driven models. Our design effectively merges the geometric clarity of optimization-based methods with the adaptive power of data-driven learning. In the forward pass, this synergy is realized by replacing the conventional photometric error with a feature-based objective. A lightweight CNN extracts robust feature maps, and the LK algorithm iteratively minimizes the difference between feature vectors at warped locations to solve for optical flow. To enable gradient flow in the backward pass, we rely on a key property from the forward optimization: once the solver converges to a solution, the gradient there is zero. We then use the implicit function theorem on this optimality condition. This allows us to directly compute how the flow changes with respect to the input CNN features. Consequently, the entire LK optimization becomes differentiable, permitting end-to-end training. Overall, the contributions of this paper include,

\begin{enumerate}[1)]
    \item A neuro‑symbolic framework that unites a data‑driven CNN feature extractor with an optimization‑based differentiable LK solver, enabling end‑to‑end trainable sparse optical flow.
    \item A fully differentiable LK layer implemented via implicit differentiation, allowing stable gradient propagation through the iterative optimizer.
    \item Effective learning from synthetic data only, which achieves robust zero‑shot generalization to real‑world scenes.
    \item Real‑time VIO integration that improves odometry accuracy, reduces tracking failures, and maintains stable performance under challenging visual conditions.
\end{enumerate}

The rest of the paper is organized as follows. Section \ref{section2} reviews related work in deep learning enhanced visual odometry, learned optical flow for VO/VIO, and differentiable optimization methods. Section \ref{section3} details our proposed method, including the system overview, descriptor-based LK optical flow estimation, backpropagation through the LK solver, and the end-to-end training procedure. Experimental setup, optical flow analysis, comparison with state-of-the-art methods, and integration into a VIO system are presented in Section \ref{section4}. Finally, Section \ref{section6} concludes the paper.

\section{RELATED WORK}
\label{section2}

%


To address the trade-off between robustness, accuracy, and efficiency in image correspondence for VO/VIO, we propose an end-to-end trainable neuro-symbolic learning system for sparse optical flow estimation. Our approach builds upon and relates to three key lines of prior work: (1) deep learning-enhanced visual odometry, (2) learned optical flow for odometry, and (3) differentiable optimization methods.

\subsection{Deep Learning Enhanced Visual Odometry}

The growing availability of both computational power and high-quality image datasets has led to considerable research into learning-based methods for visual odometry \cite{slamhand-ch13}. Among these, some replace specific components of the VO pipeline with data-driven modules \cite{xu2024airslam}, while others directly regress depth and pose from the network \cite{wang2017deepvo}. The first approach effectively utilizes mature VO frameworks and improves robustness. However, simply replacing modules in SLAM with deep learning methods often yields suboptimal results. For instance, \cite{Liu2024_VIOreport, PracticalVIO2024} point out that integrating SuperPoint \cite{superpoint} and LightGlue \cite{lightglue} with visual odometry does not always lead to improved accuracy. The second approach estimates the optimal outcome in an end-to-end manner, but is often considered to have limited generalization. This observation prompts us to consider a fundamental problem: \textit{How can we design an approach that not only harnesses the strengths of end-to-end learning but also generalizes robustly to unseen scenarios?} \par 

A recent finding shows that using learned optical flow as an intermediate representation for pose estimation delivers optimal accuracy and generalization together. DROID-SLAM \cite{teed2021droid} was the first to integrate bundle adjustment (BA) and optical flow estimation into a single network, iteratively optimizing depth and relative pose. Remarkably, it achieved strong generalization performance even when trained solely in a simulated environment. Subsequently, several state-of-the-art visual odometry systems, such as DPVO \cite{dpvo}, TartanVO \cite{tartanvo}, and MAC-VO \cite{macvo}, have adopted optical flow as an intermediate estimation result. Therefore, despite the many deep learning and SLAM combinations, we find that estimating optical flow first and then optimizing pose iteratively is a very promising path. These methods estimate dense flow through network regression, which often lacks geometric consistency. More importantly, their low efficiency limits direct use on robots.

\subsection{Learned Optical Flow for Visual Odometry}

Optical flow is a key intermediate step in visual odometry. It is commonly divided into two types: optimization-based and regression network-based methods. Optimization-based optical flow works in two steps. First, it defines an objective function based on assumptions like brightness constancy \cite{lk_flow}. Then, it solves the equation iteratively to get the flow result. Later methods improve robustness and accuracy by manually designing different objective functions. These methods provides high geometric consistency, making it suitable for direct pose estimation and widely adopted in VO/VIO systems \cite{mourikis2007msckf,qin2018vins,ov2slam}. However, relying on manual objective design, optimization methods adapt poorly to severe lighting and viewpoint changes, which holds back the development of VO systems. \par 

Regression-based optical flow estimation uses a neural network to directly predict flow results from image pairs, bypassing the need for manually defined objectives. Dense optical flow results are typically obtained directly, unlike sparse optical flow. The model learns complex motion patterns directly from training data, allowing it to handle challenging scenarios where traditional assumptions break down. This leads to strong generalization and greater robustness in practice. FlowNet \cite{flownet} is the first end-to-end trainable architecture. Building upon this, RAFT \cite{raft} introduced a framework that estimates optical flow in two stages: first by building a 4D cost volume to capture pixel-wise correlations, and then by refining the flow through iterative updates. This is an accurate and highly generalizable design, as it follows the fundamental approach of iterative optimization. Subsequent methods like GMA \cite{gma} and FlowFormer \cite{flowformer} have advanced performance by leveraging large-scale datasets and attention modules. This enables them to capture richer contextual information, thus improving the robustness and precision of flow estimation. \par 

Despite the significant strides of regression-based optical flow in accuracy and robustness, its lack of geometric consistency limits direct use in visual odometry. This is evidenced in DROID-SLAM \cite{teed2021droid}, where directly combining such flow with bundle adjustment led to low pose accuracy. To resolve this, introducing geometric constraints into the iterative estimation process is essential for improving accuracy. Later, DPVO \cite{dpvo} tried to improve the efficiency of iterative optical flow estimation, and MAC-VO \cite{macvo} incorporates optical flow estimation confidence to select geometrically consistent regions for use in Visual Odometry. Together, these findings show that network-predicted flow alone is not accurate enough—it needs geometric constraints for higher precision. We also note that traditional optimization-based flow can be used directly for pose estimation, suggesting it has higher inherent accuracy. Therefore, this work will explore \textit{whether optical flow can be accurately estimated by an end-to-end model that learns from data and satisfies precise geometric constraints}.

\begin{figure*}[!t] 
    \centering 
    \includegraphics[width=\textwidth]{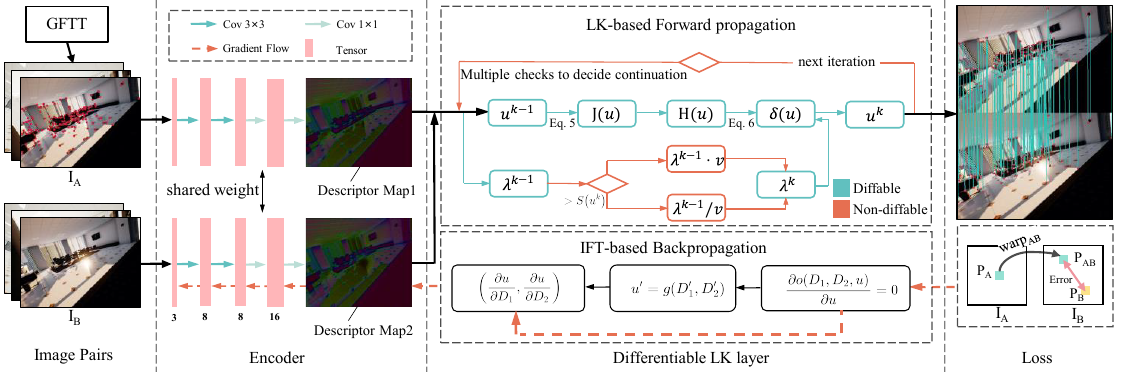}
    \caption{\textbf{The pipeline of the proposed method.} Images $\textbf{I}_A$ and $\textbf{I}_B$ are two consecutive frames in a sequence. Sparse keypoints are first detected in $\textbf{I}_A$ using the GFTT \cite{harris} detector. A shared-weight convolutional network is then used to extract descriptor maps. In the forward flow estimation, a standard LM optimization routine computes the flow from the descriptor maps, where the two orange-colored steps are non-differentiable. For backward propagation, derivatives are calculated by leveraging the optimality condition (where the gradient of the objective w.r.t. flow equals zero). The dashed orange arrows illustrate the corresponding gradient flow. The loss is computed from the error between the estimated optical flow and the ground truth.} 
    \label{architecture} 
    \end{figure*}

\subsection{Differentiable Optimization}

While implicit learning capabilities have attracted increasing attention, autonomous robotic systems still face significant challenges in achieving interpretable learning. This is especially evident in tasks involving geometric, physical, and logical learning. Overcoming these obstacles and integrating interpretable symbolic learning into data-driven models—a direction known as neuro-symbolic learning—holds strong potential to significantly enhance robotic autonomy \cite{neuro}.  \par 
Many studies have attempted to combine traditional iterative optimization algorithms with learning-based methods to achieve improved accuracy and generalization. Traditional iterative algorithms are typically non-differentiable because step sizes and iteration counts are determined by heuristic rules. Differentiable formulations are generally realized through either unrolled or implicit differentiation. The unrolled differentiation method approximates derivatives by unrolling and truncating the optimization after a fixed number of iterations, following the approach used in BA-Net\cite{tang2019ba}, DROID-SLAM \cite{teed2021droid}, and iSLAM \cite{islam}. Unlike explicit methods which track the entire optimization path, implicit differentiation uses a key shortcut: at an optimum, the derivative is zero. This allows derivatives to be calculated directly, improving both efficiency and numerical stability. This approach is widely adopted in differentiable optimization libraries such as Theseus \cite{theseus, pypose}. While these methods have enabled many end-to-end neuro-symbolic systems, to our knowledge, a differentiable formulation of the LK method for optical flow estimation has not yet been developed.

\section{METHOD}
\label{section3}

In this subsection, we systematically describe the proposed end-to-end trainable sparse optical flow method. First, we describe the unified system framework, including the network design and how it couples with the differentiable LK layer to estimate flow jointly. Next, we delve into two key components of the framework: the iterative optimization of optical flow from descriptor maps includes a forward process, along with its corresponding backward pass. Then, we introduce the end-to-end training procedure and the associated loss functions. As a final step, we implement the neuro-symbolic optical flow method within a complete VIO system.

\subsection{Overview of the System Architecture}

The proposed system, illustrated in Fig.~\ref{architecture}, comprises two parts: a neural component and a symbolic component. The former is a network encoder that extracts image descriptor maps; the latter is a differentiable LK layer that iteratively estimates optical flow from these maps. First, we introduce the network encoder in detail.

\subsubsection{Encoder Network}

The adopted network encoder shares a similar architecture with previous works such as LET-Net \cite{letnet} and SuperPoint \cite{superpoint}, with only minor differences. Unlike these methods, we do not specify an explicit definition of an optimal descriptor map; instead, it emerges implicitly from the end-to-end optical flow learning objective. Given that the required descriptor maps capture local image information rather than high-level features, the network architecture is designed to be extremely lightweight. \par 

As shown in Fig.~\ref{architecture}, the image feature encoder maps the input image $
I \in \mathbb{R}^{W \times H \times 3}$ to a descriptor map of size $W \times H \times d$. The first two convolutional layers use $3 \times 3$ kernels and expand the shared feature map to 8 channels. Then, a $1 \times 1$ convolution is applied to increase the number of channels to 16. Finally, another $1 \times 1$ convolution is used to reduce the channels of the descriptor map to $d$. A ReLU activation function~\cite{relu} is applied after each convolution, and the original image resolution is preserved throughout all convolutional operations. With the descriptor maps available, an optimization-based solver is then used to estimate the optical flow from them.

\subsubsection{Descriptor-Based LK Method}
\label{subsec:DescriptorLK}



To estimate optical flow from descriptor maps $\mathbf{D} \in \mathbb{R}^{W \times H \times d}$, we need to modify the original LK method. Specifically, the classical LK method computes optical flow by minimizing the pixel intensity difference between two frames, i.e.,
\begin{equation}
\mathbf{u}^* = \arg\min_{\mathbf{u}} \sum_{\mathbf{x}} \| I_2(\mathbf{x} + \mathbf{u}) - I_1(\mathbf{x}) \|^2,
\end{equation}
where $\mathbf{u} \in \mathbb{R}^{2}$ denotes the flow vector, $\mathbf{x} \in \mathbb{R}^{2}$ is the initial keypoints location. However, in scenarios with dynamic illumination, weak textures, or noise, the brightness constancy assumption often does not hold.  
Therefore, we modify the objective function to minimize the distance between descriptors of the two frames, i.e.,
\begin{equation}
\label{eqe:object}
\mathbf{u}^* = \arg\min_{\mathbf{u}} \sum_{\mathbf{x}} \| \mathbf{D}_2(\mathbf{x} + \mathbf{u}) - \mathbf{D}_1(\mathbf{x}) \|^2,
\end{equation}
where $\mathbf{D}_1$ and $\mathbf{D}_2$ are the descriptor maps of the first and second frames, respectively. The detailed steps to solve this optimization problem as well as the computation of its derivatives will be described in the following sections.

\subsection{Forward Pass of the Differentiable LK Layer}
\label{sec:lk}

As described in Subsection \ref{subsec:DescriptorLK}, the objective function of the LK solver is modified from minimizing pixel intensity differences to minimizing the euclidean distance between image descriptors. This constitutes a typical nonlinear optimization problem. Following prior work, we adopt the Levenberg-Marquardt (LM) method for its solution. The LM algorithm iteratively solves nonlinear least-squares problems. A key feature is that it approximates the hessian matrix using only the jacobian (first derivatives), which guides the parameter updates toward the optimum. Accordingly, we first present the computation of the first-order derivatives of the proposed descriptor-based error function, and then describe the LM iterative solution procedure, highlighting the components that are non-differentiable.

\subsubsection{First-Order Derivatives of the Descriptor-Based Error Function}

First, we define the descriptor difference between the two images as the error function, as follows
\begin{equation}
    f(\mathbf{x},\mathbf{u}) = \mathbf{D}_2(\mathbf{x} + \mathbf{u}) - \mathbf{D}_1(\mathbf{x}) .
\end{equation}
To approximate the first-order derivative of the error function, we make a second assumption — that the optical flow $ \delta \mathbf{u}$ between consecutive frames is small, i.e., $\delta \mathbf{u} \approx \mathbf{0}$.
By applying the Taylor expansion at $ \delta \mathbf{u} = 0$, we obtain,
\begin{equation}
\begin{aligned}
f(\mathbf{x} , \mathbf{u} + \delta \mathbf{u}) &= f(\mathbf{x} , \mathbf{u}) + f^{'}(\mathbf{x},\mathbf{u}) \delta \mathbf{u} \\
&= \mathbf{D}_2(\mathbf{x}+ \mathbf{u}) - \mathbf{D}_1(\mathbf{x}) + \frac{\partial \mathbf{D}_2(\mathbf{x} + \mathbf{u})}{\partial \mathbf{x} + \mathbf{u}} \delta \mathbf{u}
\end{aligned},
\end{equation}
where $\frac{\partial \mathbf{D}_2(\mathbf{x} + \mathbf{u})}{\partial \mathbf{x} + \mathbf{u}} \in \mathbb{R}^{2 \times 2}$ represents the gradient information of the descriptor of image 2 in different directions, that is,
\begin{equation}
\frac{\partial \mathbf{D}_2(\mathbf{x} + \mathbf{u})}{\partial \mathbf{x} + \mathbf{u}} = \begin{bmatrix} \nabla
\mathbf{D}_2^x(\mathbf{x} + \mathbf{u}) & 0 \\
0 & \nabla \mathbf{D}_2^y(\mathbf{x} + \mathbf{u})
\end{bmatrix}  \equiv \mathbf{J} ,
\end{equation}
where $\nabla \mathbf{D}_2^x$ represents the derivative of the descriptor map $\mathbf{D}_2$ in the $x$-direction, while $\nabla \mathbf{D}_2^y$ represents the derivative in the $y$-direction.

\subsubsection{The Iterative Levenberg-Marquardt Optimization}

\begin{algorithm}[ht]
\caption{Descriptor-Based LM Optical Flow Algorithm}
\label{alg:lm}
\begin{algorithmic}[1]
\Require Initial parameter $\mathbf{u}^{(0)}$, initial damping $\lambda^{(0)}>0$, tolerance $\varepsilon$, max iterations $K_{\max}$, factor $\nu>1$
\Ensure Estimate $\mathbf{u}^*$

\State Define $\mathbf{f}(\mathbf{u})$ and $S(\mathbf{u})=\tfrac{1}{2}\|\mathbf{f}(\mathbf{u})\|_2^2$.
\For{$k=0,1,\dots,K_{\max}-1$}
\State Compute $\mathbf{J}^{(k)}$ and $\mathbf{f}(\mathbf{u}^{(k)})$.
\State Solve Eq.~\eqref{eq:update} for $\delta\mathbf{u}$.
\State $\mathbf{u}^{\text{cand}}=\mathbf{u}^{(k)}+\delta\mathbf{u}$.
\If{$S(\mathbf{u}^{\text{cand}})<S(\mathbf{u}^{(k)})$}
\State Accept: $\mathbf{u}^{(k+1)}=\mathbf{u}^{\text{cand}}$, set $\lambda^{(k+1)}=\lambda^{(k)}/\nu$.
\Else
\State Reject: $\mathbf{u}^{(k+1)}=\mathbf{u}^{(k)}$, set $\lambda^{(k+1)}=\lambda^{(k)}\cdot\nu$.
\EndIf
\If{$\|\Delta\mathbf{u}\|_2<\varepsilon(1+\|\mathbf{u}^{(k)}\|_2)$ \textbf{or} $\|\mathbf{J}^{ (k)\top}\mathbf{f}(\mathbf{u}^{(k)})\|_\infty<\varepsilon$}
\State \textbf{break}
\EndIf
\EndFor
\State \Return $\mathbf{u}^{(k)}$
\end{algorithmic}
\end{algorithm}

After obtaining the taylor expansion of the error function, the optimal optical flow can then be solved using the LM method. 
The standard LM iteration formula is given by
\begin{equation}
\label{eq:update}
    (\mathbf{H} + \lambda \mathbf{I}) \delta \mathbf{u} = -\mathbf{J}^{\top} f(\mathbf{x} + \mathbf{u}),
\end{equation}
where $\mathbf{H} =\mathbf{J}^{\top} \mathbf{J} \in \mathbb{R}^{2\times 2}$ is the approximated hessian matrix obtained from the first-order derivatives, and $\lambda$ is an adjustable damping factor that ensures the positive definiteness of the matrix $(\mathbf{H} + \lambda \mathbf{I})$.
The classical LM optimization process is illustrated in Algorithm \ref{alg:lm}, in which two components are non-differentiable. Fig. \ref{architecture} visualizes the iterative process of the LM algorithm. Different colors distinguish between the differentiable and non-differentiable parts. This provides an intuitive illustration.

\begin{enumerate}
    \item The iterative process terminates once a predefined convergence threshold is reached. This if-else–based termination strategy makes the output non-differentiable with respect to the input.
    \item In each iteration, the algorithm updates the damping factor $\lambda$ based on the current value of the objective function. If a step fails to reduce the objective, $\lambda$ is increased; otherwise, it is decreased. This if-else decision mechanism likewise introduces non-differentiability.
\end{enumerate}
BA-Net \cite{tang2019ba} was the first to identify the sources of non-differentiability in the optimization process. It addresses the first issue by adopting a fixed number of iterations, terminating regardless of the optimization quality.
Moreover, it introduces a neural network–based update for the parameter $\lambda$, thereby avoiding the non-differentiable nature of the heuristic update process.
Although such modifications can achieve differentiability, they introduce issues such as inconsistency between training and inference. We will introduce the implicit differentiation method, which requires no modification to the original algorithm.

\subsection{Backpropagating the Differentiable LK Layer}

Before presenting the proposed method, we first provide a brief introduction to the Implicit Function Theorem to facilitate a better understanding of our approach. \par 

\subsubsection{Implicit Function Theorem (IFT)}

\begin{tm}[\cite{Krantz2012implicit}]\label{tm:ift_rewrite}
Consider a continuously differentiable mapping $f:\mathbb{R}^{n+m} \to \mathbb{R}^{m}$, 
with input $(\mathbf{a}, \mathbf{b}) \in \mathbb{R}^n \times \mathbb{R}^m$. 
Suppose there exists a point $(\mathbf{a}^{*}, \mathbf{b}^{*})$ such that
\begin{equation}
    f(\mathbf{a}^{*}, \mathbf{b}^{*}) = \mathbf{0}.
\end{equation}
Furthermore, assume that the Jacobian matrix 
$\frac{\partial f}{\partial \mathbf{b}}(\mathbf{a}^{*}, \mathbf{b}^{*})$ is invertible. 
Then, by the IFT, there exists an open set 
$V \subset \mathbb{R}^n$ containing $\mathbf{a}^{*}$ and a unique continuously differentiable function 
$g(\mathbf{a}): \mathbb{R}^n \to \mathbb{R}^m$ such that $\mathbf{b}^{*} = g(\mathbf{a}^{*})$ and
\begin{equation}
    f(\mathbf{a}{'}, g(\mathbf{a}{'})) = 0, \forall \mathbf{a}{'} \in V.
\end{equation}

Moreover, for all $\mathbf{a}' \in V$, the Implicit Function Theorem states that 
the Jacobian matrix of $g$ is given by
\begin{equation}
    \frac{\partial g}{\partial \mathbf{a}}(\mathbf{a}^{'}) = - \left[ \frac{\partial f}{\partial \mathbf{b}}(\mathbf{a}^{'}, g(\mathbf{a}^{'})) \right]^{-1} \left[ \frac{\partial f}{\partial \mathbf{a}}(\mathbf{a}^{'}, g(\mathbf{a}^{'})) \right]
\end{equation} .
\end{tm}

In practice, the IFT offers a systematic method for computing the derivative of an implicitly defined function $ g(\mathbf{a}) $ with respect to its input $\mathbf{a}$, even when $ g $ cannot be expressed in a closed-form expression. This is done by differentiating the constraint $f(\mathbf{a}, g(\mathbf{a})) = \mathbf{0}$. Solving this yields the jacobian of $g$. With the jacobian, we can perform sensitivity analysis and gradient-based optimization. This approach is key when explicit formulas are not available.

\subsubsection{Computation of Derivatives}

This method enables the learning of useful descriptors without a manually defined loss. It works by using the optical flow estimation error as a training signal to directly compute gradients for the descriptors via backpropagation. Therefore, we aim to compute
\begin{equation}
    \frac{\partial \mathbf{u}}{\partial \mathbf{D}_1} \in \mathbb{R}^{W \times H \times d \times 2 P}, \quad and \quad \frac{\partial \mathbf{u}}{\partial \mathbf{D}_2}  \in \mathbb{R}^{W \times H \times d \times 2 P},
\end{equation}
where $\mathbf{u}_i$ is the $i$-th optical flow estimate obtained as described in Sec. \ref{sec:lk}, and $\mathbf{D}_1$, $\mathbf{D}_2$ are the descriptor maps extracted by the convolutional network from images 1 and 2, respectively. However, as discussed earlier, this process is non-differentiable and cannot be expressed in a closed form. To compute the derivatives above, we first define the objective function
\begin{equation}
    o(\mathbf{D}_1, \mathbf{D}_2, \mathbf{u}) = \sum_{i=1}^{n}\|\mathbf{D}_1(\mathbf{x}_i) - \mathbf{D}_2(\mathbf{x}_i+\mathbf{u}_i)\|^2 ,
\end{equation}
where $\mathbf{x}_i$ denotes the coordinate of the $i$-th keypoint in image 1. Using the LM algorithm described in Sec. \ref{sec:lk}, the locally optimal optical flow estimate $\mathbf{u}_i^{*}$ satisfies
\begin{equation}
    \left. \frac{\partial o(\mathbf{D}_1, \mathbf{D}_2, \mathbf{u})}{\partial \mathbf{u}} \right\rvert_{\mathbf{u} = \mathbf{u}^*} = \mathbf{0} \in \mathbb{R}^{n\times 2}.
\end{equation}
This is an implicit function expression, which can be defined as $f(\mathbf{D}_1, \mathbf{D}_2, \mathbf{u})$. There exists a point $(\mathbf{D}_1, \mathbf{D}_2, \mathbf{u}^*)$ such that
\begin{equation}
    f(\mathbf{D}_1, \mathbf{D}_2, \mathbf{u}^*) = \mathbf{0},
\end{equation}
where $\mathbf{u}^*$ is the optical flow result corresponding to the local minimum obtained from $\mathbf{D}_1$ and $\mathbf{D}_2$. Hence, this function satisfies the IFT and can be differentiated implicitly. Before differentiation, we expand the function for clarity. Let $f(\mathbf{D}_1, \mathbf{D}_2, \mathbf{u}) = [f_1, \cdots, f_P]$, where $f_i \in \mathbb{R}^2$ and $i \in [1, P)$. Thus, the function can be further expanded as
\begin{equation}
    \begin{aligned}
        f_i &= \frac{\partial o(\mathbf{D}_1, \mathbf{D}_2, \mathbf{u})}{\partial \mathbf{u}_i}\\
            &= -2 (\mathbf{D}_1(\mathbf{x}_i) - \mathbf{D}_2(\mathbf{x}_i+\mathbf{u}_i))\frac{\partial \mathbf{D}_2(\mathbf{x}_i+\mathbf{u}_i)}{\partial \mathbf{u}_i}
    \end{aligned}, 
\end{equation}
where $\frac{\partial \mathbf{D}_2(\mathbf{x}_i+\mathbf{u}_i)}{\partial \mathbf{u}_i}$ denotes the derivatives of the learned descriptor map with respect to the $x$ and $y$ optical flow directions, i.e., the descriptor gradient information, which can also be represented as $\nabla \mathbf{D}_2(\mathbf{x}_i+\mathbf{u}_i)$. \par

After obtaining the explicit expression of $f(\mathbf{D}_1, \mathbf{D}_2, \mathbf{u})$, the derivatives can be computed using the IFT. According to the theorem, there exists a continuously differentiable function
\begin{equation}
    \mathbf{u}^{'} = g(\mathbf{D}_1^{'}, \mathbf{D}_2^{'}),
\end{equation}
and the partial derivatives in the neighborhood can be expressed as
\begin{equation}
    \begin{aligned}
    \frac{\partial \mathbf{u}}{\partial \mathbf{D}_1} &= -\left[\frac    {\partial f}{\partial \mathbf{u}}\right]^{-1} \left[\frac{\partial f}{\partial \mathbf{D}_1} \right] \\
    \frac{\partial \mathbf{u}}{\partial \mathbf{D}_2} &= -\left[\frac{\partial f}{\partial \mathbf{u}}\right]^{-1} \left[\frac{\partial f}{\partial \mathbf{D}_2} \right] 
   \end{aligned} ,
\end{equation}
where, since the explicit form of $f$ is known, these derivatives can be analytically solved.

\subsection{End-to-end learning with Differentiable LK Layer}

The differentiability of the LK layer is what makes end-to-end training possible. This is a key advantage over prior works that depended on manually designed loss functions to indirectly supervise descriptor learning, often leading to suboptimal performance. End-to-end training directly optimizes all components for the final objective. \par 

The proposed training pipeline (Algorithm \ref{alg:e2e-train}) enables fully differentiable, end-to-end sparse optical flow estimation. It achieves this by integrating LM optimization with implicit differentiation within a unified framework. The overall training flow is shown in Fig. \ref{architecture}. To enhance robustness, we begin by applying illumination perturbations to a pair of consecutive input frames. Harris keypoints are detected in the first frame, and their correspondences in the second are established using either ground-truth optical flow or a known geometric warp. Aimed at refining these matches, we initialize perturbed flows around them and optimize directly in a convolutional feature space using our LM-based solver. The resulting flow estimates then contribute to a robust matching loss. Crucially, for end-to-end learning, gradients are propagated through the entire LM optimization via the implicit function theorem. This allows the descriptor network and the flow solver to be trained jointly, maximizing the consistency and accuracy of the final estimates.

\begin{algorithm}[t]
\caption{End-to-End Training via Implicit Differentiation}
\label{alg:e2e-train}
\small
\begin{algorithmic}[1]
\Require Consecutive frames $\{(I_1,I_2)\}$, ground truth $\mathbf{u}^{\text{gt}}$, 
descriptor CNN $\phi_\theta$, LM solver, optimizer (e.g., Adam), 
illumination augmenter $\mathcal{A}$, Harris detector $\mathcal{H}$, 
perturbation std $\sigma$, epochs $T$, batch size $B$, keypoints per image $P$
\Ensure Trained descriptor parameters $\theta$

\vspace{2pt}
\State Define loss 
$\mathcal{L}(\hat{\mathbf{u}},\mathbf{u}^{\text{gt}})
   = \tfrac{1}{P}\sum_{i=1}^{P}\rho(\|\hat{\mathbf{u}}_i-\mathbf{u}^{\text{gt}}_i\|_2)$
and feature objective 
$o(\mathbf{D}_1,\mathbf{D}_2,\mathbf{u})
   = \sum_{i=1}^{P}\|\mathbf{D}_1(\mathbf{x}_i)
   - \mathbf{D}_2(\mathbf{x}_i+\mathbf{u}_i)\|_2^2.$

\vspace{2pt}
\For{$t=1$ \textbf{to} $T$}
  \For{each mini-batch of size $B$}
    \State Sample $B$ frame pairs $(I_1, I_2)$
    \State Apply illumination perturbation: 
      $(\tilde{I}_1,\tilde{I}_2)=\mathcal{A}(I_1,I_2)$
    \State Detect Harris keypoints 
      $\{\mathbf{x}_i\}_{i=1}^{P}\leftarrow \mathcal{H}(\tilde{I}_1)$
    \State Compute $\mathbf{x}^{\text{gt}}_i=\mathbf{x}_i+\mathbf{u}^{\text{gt}}(\mathbf{x}_i)$
    \State Initialize 
      $\mathbf{u}^{(0)}_i=(\mathbf{x}^{\text{gt}}_i-\mathbf{x}_i)+
      \varepsilon_i$, with 
      $\varepsilon_i\sim\mathcal{N}(\mathbf{0},\sigma^2\mathbf{I}_2)$
    \State Extract feature map:
      $\mathbf{D}_1=\phi_\theta(\tilde{I}_1)$, 
      $\mathbf{D}_2=\phi_\theta(\tilde{I}_2)$
    \State Estimate flow:
      $\hat{\mathbf{u}}\!\leftarrow\!
      \textbf{LM}\big(o(\mathbf{D}_1,\mathbf{D}_2,\mathbf{u}),\mathbf{u}^{(0)}\big)$
    \State Compute training loss 
      $\mathcal{L}\!=\!\mathcal{L}(\hat{\mathbf{u}},\mathbf{u}^{\text{gt}})$
      (mask invalid points)
    \Statex \textbf{Backward: Implicit Differentiation}
    \State Form $f(\mathbf{D}_1,\mathbf{D}_2,\hat{\mathbf{u}})=
      \tfrac{\partial o}{\partial \mathbf{u}}\big\rvert_{\hat{\mathbf{u}}}$,
      with Jacobian $\mathbf{J}_u=\tfrac{\partial f}{\partial \mathbf{u}}$
    \State Compute:
      $\tfrac{\partial \hat{\mathbf{u}}}{\partial \mathbf{D}_1}
      =-\mathbf{J}_u^{-1}\!\left[\tfrac{\partial f}{\partial \mathbf{D}_1}\right]$,
      \quad
      $\tfrac{\partial \hat{\mathbf{u}}}{\partial \mathbf{D}_2}
      =-\mathbf{J}_u^{-1}\!\left[\tfrac{\partial f}{\partial \mathbf{D}_2}\right]$
    \State Obtain gradient by chain rule:
      $\tfrac{\partial \mathcal{L}}{\partial \theta}
      =\big\langle\tfrac{\partial \mathcal{L}}{\partial \hat{\mathbf{u}}},
      \tfrac{\partial \hat{\mathbf{u}}}{\partial \mathbf{D}_1}
      \tfrac{\partial \mathbf{D}_1}{\partial \theta}
      +\tfrac{\partial \hat{\mathbf{u}}}{\partial \mathbf{D}_2}
      \tfrac{\partial \mathbf{D}_2}{\partial \theta}\big\rangle$
    \State Update parameters:
      $\theta \leftarrow \text{OptimizerStep}(\theta, \nabla_\theta \mathcal{L})$
  \EndFor
\EndFor
\State \Return $\theta$
\end{algorithmic}
\end{algorithm}

\subsection{VIO System with Neuro-Symbolic Optical Flow}

\begin{figure}[t]
    \centering
    \includegraphics[width=0.95\linewidth]{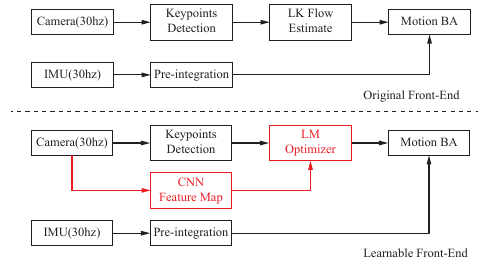}
    \caption{\textbf{Comparison between the original and the proposed learnable front-end in VINS-Fusion.} 
    The original system (top) employs a hand-crafted LK optical flow for feature tracking, 
    while the proposed version (bottom) replaces it with a learnable LM-optimized optical flow module 
    guided by a descriptor network, enabling end-to-end trainability and improved robustness.}
    \label{fig:vins-lk}
\end{figure}

As illustrated in Fig.~\ref{fig:vins-lk}, the proposed neuro-symbolic optical flow is integrated into the VINS-Fusion \cite{qin2018vins} framework. The original VINS-Fusion \cite{qin2018vins} pipeline consists of a visual front-end and an optimization-based back-end. In the visual front-end, consecutive image frames are processed to extract and track sparse keypoints using a classical LK optical flow method. These tracked keypoints are then used to estimate camera motion between frames. Simultaneously, the Inertial Measurement Unit (IMU) data are pre-integrated to provide motion priors and temporal constraints. In the back-end, both the visual measurements and the IMU pre-integration results are jointly optimized in a sliding-window pose graph to recover accurate camera poses and generate a local map.

Rather than using a manual optical flow tracker, we employ a learnable LK module that works directly on descriptor maps. Given consecutive frames, a neural network first extracts descriptor representations that are robust to illumination and texture variations. The optical flow is then refined by a differentiable LM optimizer, which uses these descriptor maps to solve a nonlinear least-squares problem, enabling precise flow estimation. The resulting flow fields provide precise inter-frame feature associations that are subsequently fused with IMU pre-integration results in the back-end. 

This integration enables the VINS-Fusion \cite{qin2018vins} system to adaptively learn descriptor representations aligned with its optimization objective, resulting in improved keypoints tracking accuracy, reduced drift, and stronger robustness to illumination and motion variations.

\section{Experiments}
\label{section4}

In this section, we provide a comprehensive description of our experimental study. We first introduce the experimental setup, including the training datasets, implementation details, and evaluation benchmarks. Next, the fundamental properties of optical flow are analyzed. The capability of the proposed method to preserve spatial sparsity while maintaining accuracy under varying motion and illumination is demonstrated. In subsection V-C, we present a comparative evaluation between our proposed sparse optical flow approach and several state-of-the-art (SOTA) methods, highlighting its superior performance in both precision and computational efficiency. Finally, in subsection V-D details the integration of our neuro-symbolic flow module into the VINS-Fusion \cite{qin2018vins} framework, showing consistent improvements in pose estimation accuracy and robustness.

\subsection{Experimental Setup}


Before presenting the experimental results, we provide a unified description of the experimental settings. These settings remain consistent across all subsequent experiments to avoid repetitive explanations. They mainly include an introduction to the training data, details and parameters of the network training, descriptions of the datasets used for inter-frame relative pose estimation, the datasets and hardware used in the odometry experiments, as well as information about the computing devices employed.

\subsubsection{Training data}
We adopt the TartanAir dataset \cite{tartanair} for training our sparse optical flow network. TartanAir is a large-scale, photo-realistic dataset generated using the AirSim simulation platform, providing diverse indoor and outdoor environments with ground-truth depth, optical flow, and camera pose information. The dataset features varied lighting conditions, weather effects, and motion patterns, enabling our network to learn robust representations against illumination changes and texture ambiguities.

\subsubsection{Training details}
The proposed network is implemented in \textit{PyTorch} and trained on the TartanAir dataset using color image pairs of resolution $640 \times 480$. Each batch contains 1 sample, and up to 200 salient keypoints are detected per image based on corner strength. The differentiable LK layer is unrolled for up to 50 iterations within a window size of $11 \times 11$. The model was trained on an NVIDIA RTX 4090 GPU for approximately 1 day to obtain the final network weights. 

Optimization is performed using the \textit{Adam} optimizer with a learning rate of $1\times10^{-3}$ for 1000 epochs. The model with the lowest training loss is saved as the final checkpoint for subsequent evaluations.

\subsubsection{Evaluation Datasets}  
For evaluation of both inter-frame correspondence and odometry performance, we employ a diverse set of six publicly available benchmarks covering different scenes and motion/illumination regimes.

\textbf{Optical flow evaluation:}  
\begin{itemize}  
  \item \textbf{HPatches} \cite{Balntas2017HPatches} – The Homography-Patches dataset comprises 116 image sequences of 6 frames each, with controlled illumination changes (57 sequences) and viewpoint changes (59 sequences). It provides ground-truth homographies for matching and descriptor benchmarking. 
  \item \textbf{TartanAir} \cite{tartanair} – A large-scale, photo-realistic synthetic dataset collected in Unreal Engine environments, providing RGB (and other modalities) image sequences with ground-truth poses, depths, optical flow and challenging lighting, weather and motion patterns. 
\end{itemize}  

\textbf{Odometry / visual-inertial benchmark datasets:}  
\begin{enumerate}  
  \item \textbf{EuRoC MAV} \cite{Burri2016EuRoC} – A visual-inertial dataset recorded from a micro aerial vehicle in indoor environments, containing synchronized stereo images, IMU data and accurate ground-truth trajectories across sequences of varying difficulty.  
  \item \textbf{KITTI Odometry} \cite{Geiger2012KITTI} – A real-world automotive dataset recorded in urban and rural driving environments, with stereo images, LiDAR/IMU/GPS sensors and ground-truth poses for odometry benchmarking.
  \item \textbf{UMA-VI} \cite{ZuñigaNoel2020UMA_VI} – A handheld visual-inertial dataset captured in indoor/outdoor mixed scenarios with challenging lighting and low-texture conditions, over 80 minutes of data, suitable for VIO algorithm evaluation.
  \item \textbf{AQUALOC} \cite{Ferrera2019AQUALOC} – An underwater visual-inertial-pressure dataset for localization near the seabed, recorded with a monocular camera, IMU and pressure sensor in harbor and deep-water archaeological sites – extending evaluation to extreme environments.
\end{enumerate}

By covering synthetic and real domains, aerial and ground vehicles, high-texture and low-texture scenes, and even underwater settings, this collection allows us to comprehensively assess the robustness, generalization and domain transfer capability of our proposed method.

\subsection{Analysis of Optical Flow Properties}
Since the proposed method builds upon the traditional LK optical flow by incorporating end-to-end learning, a natural question arises: why does this combination lead to better performance, and how does it achieve such improvement? To address this, we first conduct an in-depth analysis of several fundamental properties of the proposed optical flow method. \par 

\subsubsection{Trackability Analysis}
\begin{figure}[t]
    \centering
    \includegraphics[width=0.95\linewidth]{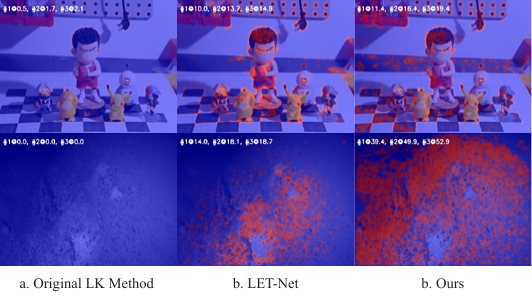}
    \caption{\textbf{Comparison of trackable regions among different methods.}  Red areas indicate regions that are easy to track, while blue areas denote low-texture or ambiguous regions that are difficult to track. 
    Compared with the Original LK Method (a) and LET-Net (b), our approach (c) produces a broader and more consistent distribution of trackable regions, 
    demonstrating stronger robustness under varying illumination and surface texture conditions.}
    \label{fig:reg}
\end{figure}

\begin{table}[t]
\centering
\footnotesize
\setlength{\tabcolsep}{3.5pt}
\caption{Median proportion (\%) of trackable regions under different pixel error thresholds.
Each cell shows the results for $<$1\,px / $<$2\,px / $<$3\,px.}
\label{tab:trackable_region_final}
\begin{tabular}{lccc}
\toprule
\multirow{2}{*}{Seq.} & \multicolumn{3}{c}{Trackable Regions @1/@2/@3 $\uparrow$} \\
\cmidrule(lr){2-2} \cmidrule(lr){3-3} \cmidrule(l){4-4}
 & LK Method & LET-Net \cite{letnet} & Ours \\
\midrule
MH000 & 0.02 / 0.05 / 0.08 & 6.78 / 8.31 / 8.69 & \textbf{19.90 / 24.48 / 27.12} \\
MH001 & 0.35 / 0.79 / 1.03 & 13.70 / 17.43 / 18.48 & \textbf{13.29 / 17.76 / 20.41} \\
MH002 & 0.01 / 0.02 / 0.02 & 10.79 / 12.38 / 12.64 & \textbf{17.08 / 20.03 / 22.04} \\
MH003 & 0.12 / 0.21 / 0.25 & 10.93 / 13.10 / 13.58 & \textbf{26.11 / 31.04 / 33.90} \\
MH004 & 0.02 / 0.09 / 0.13 & 2.60 / 3.65 / 3.95 & \textbf{3.55 / 4.75 / 5.51} \\
MH005 & 0.83 / 1.40 / 1.58 & 8.17 / 11.95 / 12.71 & \textbf{11.89 / 15.81 / 17.91} \\
MH006 & 0.04 / 0.12 / 0.17 & 11.05 / 13.33 / 13.89 & \textbf{25.17 / 29.78 / 32.69} \\
MH007 & 0.09 / 0.19 / 0.26 & 14.81 / 17.34 / 18.00 & \textbf{24.72 / 29.82 / 32.72} \\
\bottomrule
\end{tabular}
\end{table}

We evaluate the trackability of different algorithms by measuring the size of easily trackable regions within an image. Specifically, all pixels are selected as keypoints, and random perturbations of 15 pixels are added to their positions. The proposed optical flow algorithm is then used to recover the true correspondences from these noisy initializations. If the correct correspondence can be accurately recovered, the location is considered trackable; otherwise, it is regarded as non-trackable. Eight test sequences from the TartanAir dataset are used for this experiment, which are different from those employed during training. \par 

This analysis evaluates the spatial distribution and robustness of trackable regions generated by different optical flow methods. We measure the proportion of pixels whose tracking errors remain below specific thresholds (e.g., $<$1, $<$2, $<$3 pixels), reflecting how well each algorithm maintains stable feature correspondences under varying texture and illumination conditions. As shown in Fig.~\ref{fig:reg} and Table~\ref{tab:trackable_region_final}, our learnable LK approach produces a broader and more consistent range of trackable regions, especially in low-texture or partially illuminated areas. This indicates stronger feature discrimination and better local convergence behavior compared to classical LK and descriptor-based methods. \par 

\begin{figure}[t]
    \centering
    \includegraphics[width=0.95\linewidth]{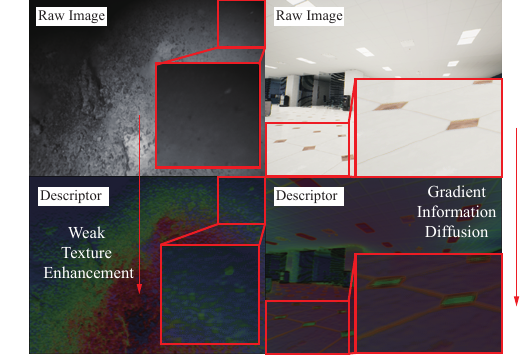}
    \caption{\textbf{Descriptor map visualization}. The top two images are the original images, and the bottom two are the descriptor maps, corresponding to an indoor scene and an underwater weak-texture environment. From the magnified regions, it can be observed that the proposed method enhances the nearly textureless areas of the original images, strengthens the image gradient information, and propagates these gradients to the surrounding regions.}
    \label{fig:feature_visualization}
\end{figure}

A deeper question concerns why the proposed method performs better and how the network collaborates with the optimization algorithm to estimate optical flow. To explore the possible reasons, we rescale the network outputs to the range of 0--255 and visualize them as RGB images, as shown in Fig.~\ref{fig:feature_visualization}. From the visualization, it can be observed that the network primarily performs three tasks: 
\begin{enumerate}
    \item extracting edge information, including both intensity and orientation; 
    \item enhancing weakly textured regions; 
    \item gradually propagating edge cues to surrounding areas, 
          which likely improves convergence under large viewpoint changes. 
\end{enumerate}
These functionalities are automatically learned through end-to-end training 
rather than manually designed, thereby enhancing the robustness of optical flow estimation 
under varying illumination, weak textures, and large perspective variations.

\subsubsection{Convergence Behavior Analysis}

\begin{figure}[t]
    \centering
    \includegraphics[width=0.95\linewidth]{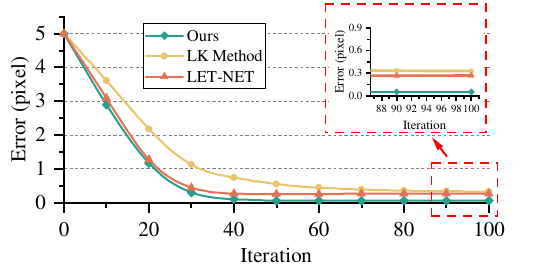}
    \caption{\textbf{Optical flow error curves over iterations.} The plot compares the convergence behavior of different methods (LK Method \cite{lk_flow}, LET-NET \cite{letnet}, and Ours). The proposed method achieves faster error reduction and lower final error, demonstrating better stability and accuracy during iterative refinement.}
    \label{fig:convergence_error}
\end{figure}

Our end-to-end training approach means that the proposed network not only extracts illumination-robust features such as edge information but is also optimized specifically for the LK iterative optical flow process. Therefore, for the same optical flow estimation, our method is expected to achieve faster convergence and smaller error. To verify this, we conducted experiments on the TartanAir test dataset, setting the initial optical flow error to 5 pixels. \par 

As shown in Fig.~\ref{fig:convergence_error}, the convergence behavior of the proposed method is compared with the traditional LK \cite{lk_flow} Method and the LET-NET \cite{letnet} model. Our method achieves faster error reduction and reaches a lower steady-state error compared to the other two. In the experiments, we found that the proposed method actively enhances weak-texture regions and propagates gradient information to their surrounding areas. This not only facilitates better optimization during the iterative process but also significantly reduces tracking errors in low-texture regions. As a result, both the convergence speed and accuracy are improved simultaneously.

\subsubsection{Inference Efficiency Analysis}
\begin{table}[t]
        \begin{center}
                \caption{INFERENCE EFFICIENCY COMPARISON}
                \label{inference}
                \setlength{\tabcolsep}{2mm}{
                \begin{tabular}{ccccc}
                \toprule
                \multirow{2}{*}{\textbf{Method}} & \multirow{2}{*}{\textbf{Params/M} $\downarrow$} &
                \multirow{2}{*}{\textbf{GFLOPs} $\downarrow$} & 
                \multicolumn{2}{c}{Interface time (ms) $\downarrow$} \\
                \cmidrule{4-5} 
                ~ & ~ & ~ &
                \textbf{GPU}  &
                \textbf{CPU} \\
                \midrule
                SuperPoint \cite{superpoint} & 22.23 & 1.30 & 8.22 & 114.72\\
                ALIKE-T\cite{alike} & 1.82 & 0.83 & 49.95 & 94.81 \\
                DISK \cite{disk} & 98.97 & 1.09 & 116.36 & 589.78 \\
                XFeat \cite{xfeat} & 1.10 & 0.64 & 4.26 & 11.17 \\
                Ours & 1.01e-3 & 0.27 & 0.84 & 5.21 \\ 
            \bottomrule
        \end{tabular}}
    \end{center}
\end{table}

\begin{table*}[h]
    \begin{center}
        \caption{evaluation results of feature matching}
        \label{fund}
        \setlength{\tabcolsep}{1.7mm}{
        \begin{tabularx}{\textwidth}{lcccccccc|cccccccc}
            \toprule
             ~ & \multicolumn{8}{c}{\%Inlier $\uparrow$}  & \multicolumn{8}{c} {Error$\downarrow$}  \\ 
            \cmidrule{2-17} 
            ~ & H00 & H01 & H02 & H03 & H04 & H05 &H06 & H07 & H00 & H01 & H02 & H03 & H04 & H05 & H06 & H07  \\ 
            \cmidrule{2-17} 
            RAFT\cite{raft} & 23.3 & 17.9 & 17.1 & 12.1 & 11.9 & 20.2 & 15.7 & 22.3 & 12.02 & 15.48 & 16.29 & 25.81 & 31.46 & 15.33 & 28.17 & 15.82 \\
            SEA-RAFT(M)\cite{sea_raft} & 20.6 & 16.0 & 14.8 & 10.7 & 12.4 & 17.3 & 17.4 & 22.1 & 15.30 & 16.28 & 19.55 & 19.89 & 19.99 & 21.16 & 16.46 & 14.22 \\
            SEA-RAFT(S)\cite{sea_raft} & 21.1 & 16.0 & 16.9 & 11.9 & 11.4 & 20.5 & 17.8 & 22.9 & 14.45 & 16.19 & 16.06 & 17.81 & 22.13 & 16.63 & 16.21 & 12.71 \\
            Key.Net(MS)\cite{keynet} & 25.5 & 20.0 & 23.4 & 16.8 & 14.6 & 26.8 & 20.0 & 25.3 & 10.14 & 11.74 & 9.47 & 13.71 & 23.19 & 7.89 & 12.62 & 10.52 \\
            ALIKE-T \cite{alike} & 25.3 & 22.6 & 24.0 & 16.5 & 14.9 & 26.3 & 19.6 & 25.6 & 9.58 & 10.94 & 9.33 & 13.95 & 21.23 & 7.90 & 12.90 & 10.47 \\
            SuperPoint \cite{superpoint} & 26.1 & 20.3 & 23.6 & 17.5 & 14.1 & 26.7 & 20.2 & 25.4 & 9.36 & 11.50 & 9.35 & 13.41 & 21.93 & 8.00 & 12.53 & 10.49 \\
            DISK\cite{disk} & \textcolor{red}{29.4} & \textcolor{red}{27.6} & 24.5 & 21.1 & 14.8 & \textcolor{red}{28.4} & 20.0 & 26.8 & 9.66 & \textcolor{red}{9.01} & 9.23 & 12.14 & 19.72 & \textcolor{red}{7.18} & 12.59 & 9.49 \\
            XFeat \cite{xfeat} & 25.8 & 21.0 & 23.4 & 16.1 & 14.4 & 26.5 & 19.3 & 25.4 & 9.59 & 11.85 & 9.77 & 14.33 & 21.8 & 7.89 & 13.42 & 10.82 \\
            SFD2 \cite{sfd2} & 26.7 & 21.6 & 24.1 & 17.3 & 15.4 & 26.6 & 20.0 & 25.5 & 9.92 & 11.10 & 9.18 & 13.38 & 20.91 & 7.83 & 12.54 & 10.42 \\
            R2D2 \cite{r2d2} & 27.2 & 21.1 & 23.3 & 17.1 & 14.8 & 27.2 & 20.2 & 26.8
            & 9.21 & 11.10 & 9.53 & 13.19 & 21.61 & 7.84 & 12.60 &  9.48 \\
            DISK\cite{disk}+LightGlue\cite{lightglue} & \textcolor{green}{29.2} & \textcolor{green}{26.5} & 21.9 & \textcolor{green}{21.9} & 12.9 & 27.2 & 18.6 & 18.6
            & 9.19 & 9.74 & 11.73 & 11.15 & 22.21 & 8.64 & 16.34 & 16.34 \\
            SuperPoint\cite{superpoint}+LightGlue\cite{lightglue} & 23.8 & 17.3 & 18.9 & 13.3 & 12.7 & 19.4 & 17.5 & 23.3
            & 10.84 & 15.98 & 12.85 & 17.82 & 21.75 & 16.66 & 18.15 & 12.61 \\
            Harris \cite{harris} + LK \cite{lk_flow} & 26.9 & 21.2 & 23.8 & 16.8 & 14.9 & 27.3 & 19.5 & 25.3 & 9.58 & 11.38 & 9.80 & 13.61 & 21.65 & 7.78 & 12.84 & 10.62 \\
            LET-NET \cite{letnet} & 28.1 & 21.6 & \textcolor{green}{26.7} & 20.1 & \textcolor{green}{16.8} & \textcolor{green}{28.2} & \textcolor{green}{21.9} & \textcolor{green}{27.9} & \textcolor{green}{8.71} & 11.00 & \textcolor{green}{7.95} & \textcolor{green}{10.65} & \textcolor{green}{14.65} & \textcolor{red}{7.18} & \textcolor{green}{10.15} & \textcolor{green}{9.21} \\
            Ours  & 28.9 & 23.0 & \textcolor{red}{27.4} & \textcolor{red}{21.2} & \textcolor{red}{18.2} & 27.9 & \textcolor{red}{23.0} & \textcolor{red}{28.4} & \textcolor{red}{8.17} & \textcolor{green}{9.51} & \textcolor{red}{7.75} & \textcolor{red}{9.72} & \textcolor{red}{12.00} & 7.39 & \textcolor{red}{9.19} & \textcolor{red}{8.82} \\
            \bottomrule
        \end{tabularx}}
    \end{center}
\end{table*} 

The proposed sparse optical flow model is highly lightweight and can be efficiently deployed on mobile CPU or GPU devices.
As shown in Table~\ref{inference}, the proposed method achieves the best inference efficiency among all compared networks. With only $1.01\times10^{-3}$M parameters and 0.27 GFLOPs, it is much lighter than existing methods such as SuperPoint \cite{superpoint} and XFeat \cite{xfeat}. The inference time for 320×240 images reaches 0.84 ms on GPU and 5.21 ms on CPU, showing clear advantages in speed and computational cost. These results demonstrate that our method can run in real time even on low-power devices. \par 

Classical end-to-end optical flow methods are typically composed of an encoder and an iterative refinement module, making their computational efficiency highly dependent on the image matching difficulty. This paradigm differs significantly from our proposed approach; hence, a direct efficiency comparison is not meaningful. For an intuitive reference, we take the classical RAFT \cite{raft} model as an example — it contains tens of millions of parameters and requires over 10 GFLOPs per frame. In contrast, our method is extremely lightweight, with minimal computational cost and memory footprint, making it highly efficient and easy to deploy on various devices.

\subsection{Comparison with State-of-the-Art Methods}
To further assess the advantages of the proposed image matching method in terms of accuracy and efficiency, comparisons are made with existing state-of-the-art learning-based optical flow, handcrafted optical flow, and learned sparse matching methods. Specifically, we first compare the classical image matching methods commonly used in SLAM on the HPatches \cite{Balntas2017HPatches} dataset, demonstrating the robustness of the proposed method to changes in viewpoint and illumination. Then, the fundamental matrix estimation error is used to compare the latest learning-based optical flow and learned sparse matching methods on the test sequences of the TartanAir \cite{tartanair} dataset. Finally, to visually demonstrate the advantages of the proposed method in VO applications, a very simple visualization result is presented. \par

\subsubsection{Tracking Error}
\begin{figure}[!t]
        \centering
        \includegraphics[width=0.48\textwidth]{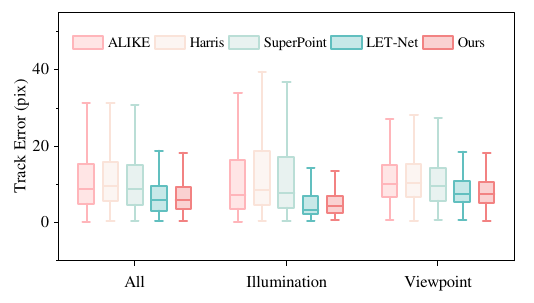}
        \caption{\textbf{Boxplot of optical flow tracking matching results.} The validation was carried out in illumination, viewpoint, and all images, respectively, based on the HPatches \cite{Balntas2017HPatches} dataset. The smaller tracking error means that the proposed features are easier to track. The proposed method achieves lower error results in both illumination and viewpoint varying datasets.} 
        \label{fig_track} 
        \end{figure}

In this experiment, we compared the robustness of common image matching methods in classical VO systems to changes in viewpoint and illumination. In the experiment, 1000 keypoints were first extracted from the first frame with a NMS size of 6. Then, the reference positions of the keypoints in the second frame were calculated using the ground-truth homography matrix provided by the dataset. Both the ALIKE \cite{alike} and SuperPoint \cite{superpoint} methods extracted learned descriptors and computed the matching relationship between the two frames using brute-force matching, while the classic Harris \cite{harris} keypoints used LK optical flow to obtain the matching relationship. The error statistics obtained by calculating the absolute matching errors of all keypoints are shown in Fig. \ref{fig_track}. \par 

The experimental results indicate that our method performs slightly better than LET-NET \cite{letnet} and clearly outperforms other methods, whether in image pairs with viewpoint changes or illumination changes. This is because both LET-NET \cite{letnet} and our method use the same combination of learning and symbolic approaches. Additionally, the HPatches dataset consists of texture-rich and clear data, making it difficult to clearly show the differences between the two methods.

\subsubsection{Fundamental Matrix Error}
Optical flow tracking error is the most intuitive measure, yet it is not entirely accurate, as the goal of this paper is applications such as visual odometry. A simpler approach is to evaluate the performance of the proposed method through the accuracy of the fundamental matrix estimation between consecutive frames. Therefore, we used test sequences from the TartanAir \cite{tartanair} dataset to compare the accuracy of the latest optical flow estimation methods, sparse image matching, and other approaches in the fundamental matrix estimation task. \par 

In the experiment, two adjacent frames of the image sequence are considered as a pair. A maximum of 1000 keypoints are extracted from the previous frame, with NMS set to 6. For dense optical flow estimation algorithms such as RAFT \cite{raft} and SEA-RAFT \cite{sea_raft}, we used the classic Harris \cite{harris} keypoints for a unified evaluation. For image matching methods based on descriptor matching, we also used brute-force matching to obtain the association results between the two frames. Additionally, some of the latest graph convolution network-based methods, such as LightGlue \cite{lightglue}, are capable of directly obtaining the matching results. \par 


Since directly calculating the error between the fundamental matrices is challenging, the inlier rate of keypoints and the average distance of features to the epipolar line are used to represent the fundamental matrix error, as referenced in \cite{evo_fund}. The keypoints in the first frame of the image are projected onto the second image using the fundamental matrix, which results in a straight line, called the epipolar line. According to the geometric projection relationship, the correctly tracked positions in the second frame will lie on this epipolar line. Therefore, keypoints whose vertical distance to the epipolar line is less than a threshold are called inliers. The percentage of inliers among all keypoints in the first frame is referred to as the inlier rate. Additionally, the mean of all the vertical distances between inliers and the epipolar line is calculated as the fundamental matrix error. \par 

Table \ref{fund} shows the inlier rate and error results, including a variety of the latest image matching methods. Among them, SEA-RAFT \cite{sea_raft} and RAFT \cite{raft} represent the latest methods that directly obtain dense optical flow results through the network. Both of these methods show significantly lower inlier counts and higher errors compared to the proposed method. We believe that the difficulty of optical flow estimation between consecutive frames is lower, and our method achieves higher accuracy in such cases. We will elaborate on this point further in the next experiment. \par 

Additionally, methods like ALIKE \cite{alike}, SuperPoint \cite{superpoint}, and DISK \cite{disk} represent approaches that use learning-based keypoint and descriptor matching trained in different ways. These methods perform well on certain sequences but are outperformed by the proposed method on most sequences. We believe this is because they typically do not account for the continuity between images and rely solely on descriptor matching. Learning-based keypoint methods and LightGlue \cite{lightglue} represent the current state-of-the-art image matching methods, which have been tried in VO systems. However, they  do not significantly improve the fundamental matrix results. They are better suited for solving large baseline image matching problems in 3D reconstruction and do not improve estimation accuracy in small baseline images. \par 

Therefore, the proposed method performs better in the fundamental matrix estimation task across the majority of sequences between consecutive frames and significantly outperforms non-end-to-end methods with similar ideas (such as LET-NET\cite{letnet}).

\subsubsection{Visualization Experiment}

\begin{figure}[!t]
        \centering
        \includegraphics[width=0.48\textwidth]{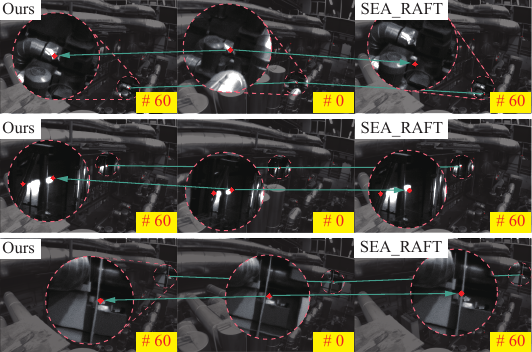}
        \caption{\textbf{Schematic diagram of the position change of sparse keypoints in continuous tracking.} The middle image shows the initial keypoint positions. The left and right images respectively display the keypoint positions after 60 consecutive tracking iterations using our method and the latest optical flow regression estimation algorithm, SEA\_RAFT \cite{sea_raft}. The sparse keypoints estimated by SEA\_RAFT \cite{sea_raft} exhibit noticeable drift (as shown in the zoomed-in images). In contrast, the proposed method maintains keypoint accuracy even after 60 tracking iterations.} 
        \label{fig_track_error} 
        \end{figure}

End-to-end learned dense optical flow often uses larger training datasets and more computational resources. However, it performs poorly in our experiments, which may seem puzzling. To address this, we aim to demonstrate the key advantage of the proposed sparse optical flow through a simple and intuitive small experiment. \par 

We believe that regression-based optical flow estimation methods like RAFT \cite{raft} are well-suited for handling difficult flow estimation tasks but may struggle with simpler tasks, where optimization-based methods like LK perform better. On the other hand, end-to-end dense methods aim to obtain an optical flow field that minimizes the global error for the entire image. However, the per-pixel accuracy of such methods is often insufficient. It is precisely this pixel-level accuracy that is of greater concern for VO tasks. To test this, we conducted a small experiment using the real EuRoC \cite{Burri2016EuRoC} dataset, randomly selecting 60 consecutive frames. In the first frame, 20 keypoints were extracted as the initial positions. Then, the optical flow between each pair of frames was estimated step by step, transferring the keypoints from the first frame to the last (the 60th frame). This process simulates a real VO scenario and amplifies the optical flow tracking errors through cumulative mistakes, making them easier to observe. \par

As shown in Fig. \ref{fig_track_error}, we visually present the cumulative optical flow error results. The three enlarged boxes in the figure show a comparison between our method and SEA\_RAFT \cite{sea_raft}. It can be observed that the proposed method, which combines end to end learning and optimization-based solutions, maintains higher accuracy over multiple tracking steps, while methods that directly obtain optical flow results from the network do not perform as well. In VO applications, even small optical flow errors in simple scenes can lead to the gradual accumulation of pose estimation errors. This result more intuitively highlights the accuracy difference between the two methods and explains why the proposed method performs better. Therefore, the proposed sparse optical flow can be directly integrated into VO systems, whereas DROID \cite{teed2021droid} requires multiple iterations to estimate the pose.

\subsection{VIO Evaluation on Public Datasets}

The proposed optical flow is designed for VO/VIO tasks and has been integrated into the classical VINS-Fusion \cite{qin2018vins} framework. Therefore, evaluating the performance of the proposed VIO system against state-of-the-art VIO algorithms is essential. To validate its robustness under dynamic lighting and weak-texture conditions, we conducted extensive experiments on the enhanced EuRoC dataset \cite{Burri2016EuRoC}, the dynamic-lighting UMA-VI \cite{ZuñigaNoel2020UMA_VI} dataset, and a weak-texture underwater dataset \cite{Ferrera2019AQUALOC}. \par 

Following most VO/VIO papers, we use the ATE (Absolute Trajectory Error) between the estimated and ground-truth trajectories as the evaluation metric. ATE measures the global consistency of the entire trajectory and is defined as the root mean square of the position errors between the estimated and ground-truth trajectories at corresponding timestamps after optimal rigid alignment. \par

\subsubsection{Experiments on the Enhanced EuRoC Dataset}
\begin{figure}[!t]
        \centering
        \includegraphics[width=0.48\textwidth]{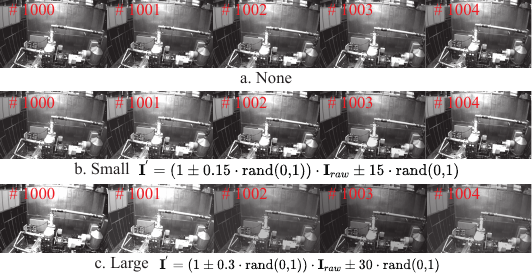}
        \caption{\textbf{Illumination augmentation on the EuRoC dataset.} The last row shows the sequences with large illumination disturbances, the middle row shows those with small disturbances, and the bottom row shows the original sequences. The illumination perturbations are generated by applying absolute and relative pixel transformations to the images using two random variables.} 
        \label{fig_enhance} 
\end{figure}
        
\begin{table*}[t]
\centering
\caption{Comparison of mean/median ATE (m) on EuRoC sequences under different illumination disturbances.
All values are in meters (m). Each cell shows mean/median ATE, and the best result is in \textbf{bold}.}
\label{tab:euroc_full}
\setlength{\tabcolsep}{2.2mm}{
\begin{tabular}{cccccccccccc}
\toprule
\multirow{2}{*}{\textbf{Seq.}} &
\multicolumn{3}{c}{\textbf{None (m)}} & &
\multicolumn{3}{c}{\textbf{Small (m)}} & &
\multicolumn{3}{c}{\textbf{Large (m)}} \\
\cmidrule{2-4} \cmidrule{6-8} \cmidrule{10-12}
~ & VINS\cite{qin2018vins} & LET-VINS\cite{letnet} & Ours & & VINS\cite{qin2018vins} & LET-VINS\cite{letnet} & Ours & & VINS\cite{qin2018vins}  & LET-VINS\cite{letnet} & Ours \\
\midrule
MH\_01 & 0.24/0.26 & 0.26/0.27 & \textbf{0.22/0.24} & &
0.27/0.26 & 0.25/0.24 & \textbf{0.25/0.25} & &
0.51/0.49 & 0.48/0.45 & \textbf{0.39/0.37} \\

MH\_02 & 0.17/0.17 & 0.18/0.18 & \textbf{0.17/0.17} & &
0.18/0.16 & 0.21/0.23 & \textbf{0.17/0.16} & &
0.48/0.47 & 0.41/0.39 & \textbf{0.32/0.30} \\

MH\_03 & 0.24/0.22 & \textbf{0.22/0.22} & 0.23/0.23 & &
0.25/0.23 & 0.27/0.25 & \textbf{0.23/0.25} & &
1.27/1.13 & 0.38/0.38 & \textbf{0.26/0.26} \\

MH\_04 & 0.42/0.43 & \textbf{0.35/0.29} & 0.42/0.44 & &
0.57/0.53 & 0.50/0.48 & \textbf{0.41/0.41} & &
\textcolor{red}{\textbf{F}} & 0.81/0.80 & \textbf{0.45/0.46} \\

MH\_05 & 0.28/0.25 & \textbf{0.29/0.26} & 0.28/0.24 & &
0.30/0.26 & 0.35/0.30 & \textbf{0.30/0.28} & &
\textcolor{red}{\textbf{F}} & 0.67/0.63 & \textbf{0.35/0.28} \\

V1\_01 & 0.10/0.11 & \textbf{0.10/0.11} & 0.11/0.11 & &
0.13/0.13 & 0.13/0.13 & \textbf{0.11/0.11} & &
0.25/0.23 & 0.24/0.22 & \textbf{0.14/0.14} \\

V1\_02 & 0.10/0.10 & 0.10/0.10 & \textbf{0.09/0.08} & &
0.10/0.09 & 0.10/0.09 & \textbf{0.08/0.08} & &
\textcolor{red}{\textbf{F}} & 0.46/0.45 & \textbf{0.09/0.08} \\

V1\_03 & 0.10/0.09 & 0.11/0.10 & \textbf{0.08/0.06} & &
0.12/0.11 & 0.10/0.10 & \textbf{0.08/0.08} & &
\textcolor{red}{\textbf{F}} & 0.58/0.47 & \textbf{0.40/0.34} \\

V2\_01 & 0.09/0.07 & 0.10/0.08 & \textbf{0.10/0.07} & &
0.10/0.08 & 0.08/0.07 & \textbf{0.11/0.08} & &
\textcolor{red}{\textbf{F}} & 0.64/0.61 & \textbf{0.11/0.10} \\

V2\_02 & 0.10/0.09 & 0.10/0.08 & \textbf{0.07/0.06} & &
0.11/0.08 & 0.12/0.11 & \textbf{0.10/0.08} & &
1.02/0.99 & 0.48/0.44 & \textbf{0.10/0.08} \\

V2\_03 & \textbf{0.26/0.27} & 0.28/0.30 & 0.39/0.42 & &
0.37/0.35 & 0.32/0.32 & \textbf{0.24/0.24} & &
5.78/5.74 & \textcolor{red}{\textbf{F}} & \textbf{1.57/1.37} \\
\bottomrule
\end{tabular}}
\vspace{-2mm}
\end{table*}

\begin{table*}[t]
\centering
\caption{Comparison of mean/median ATE (m) on UMA-VI sequences under different illumination and environmental conditions.
All values are in meters (m). Each cell shows mean/median ATE, the best result is marked in \textcolor{red}{red}, and the second-best result is marked in \textcolor{green}{green}.}
\label{tab:uma}
\setlength{\tabcolsep}{2.2mm}{
\begin{tabular}{ccccccccc}
\toprule
\textbf{Sequence} & \textbf{PL-SLAM} \cite{plslam} & \textbf{ORB-SLAM3}\cite{orbslam3}  & \textbf{Basalt}\cite{basalt} & \textbf{OKVIS} \cite{okvis} & \textbf{DROID} \cite{teed2021droid} & \textbf{AirSLAM} \cite{xu2024airslam} &\textbf{VINS} \cite{qin2018vins} & \textbf{Ours} \\
\midrule
two-floors-csc1 & \textcolor{red}{\textbf{F}} & \textcolor{red}{\textbf{F}} & 0.760 & 0.154 & 0.341 & \textcolor{red}{0.066} & 0.335  & \textcolor{green}{0.150} \\
two-floors-csc2 & \textcolor{red}{\textbf{F}} & \textcolor{red}{\textbf{F}} & 1.211 & 0.679 & \textcolor{green}{0.299} & \textcolor{red}{0.190} & 1.108 &  0.700 \\
third-floor-csc1 & 4.478 & 0.863 & 0.420 & 0.287 & 0.048 & 0.070 & \textcolor{green}{0.047} &  \textcolor{red}{0.047} \\
third-floor-csc2 &6.068 & 0.149 & 0.590 & 0.271 & 0.890 & 0.127 & \textcolor{green}{0.106} & \textcolor{red}{0.068} \\
long-walk-eng & \textcolor{red}{\textbf{F}} & \textcolor{red}{\textbf{F}} & 5.046 & 3.005 & \textcolor{red}{\textbf{F}} & \textcolor{green}{1.801} & 2.382 &  \textcolor{red}{0.353} \\
lab-module-csc & \textcolor{red}{\textbf{F}} & \textcolor{red}{\textbf{F}} & 0.403 & 0.579 & \textcolor{green}{0.319} & 0.979 & 0.570 &  \textcolor{red}{0.173} \\
lab-module-csc-rev & \textcolor{red}{\textbf{F}} & \textcolor{red}{0.063} & 0.486 & 0.861 & 0.364 & 0.504 &  \textcolor{green}{0.302} & 0.399 \\
conference-csc1 & 2.697 & \textcolor{red}{\textbf{F}} & 1.270 & 1.118 & 0.711 & \textcolor{green}{0.490} & 0.494 & \textcolor{red}{0.479} \\
conference-csc2 & 1.596 & \textcolor{red}{\textbf{F}} & 0.682 & 0.470 & 0.135 & \textcolor{red}{0.091} & 0.283 & \textcolor{green}{0.106} \\
conference-csc3 & \textcolor{red}{\textbf{F}} & 0.426 & 0.469 & 0.088 & 0.724 & 0.088 & \textcolor{red}{0.042} & \textcolor{green}{0.044} \\
\bottomrule
\end{tabular}}
\vspace{-2mm}
\end{table*}

EuRoC is a classical dataset widely used for evaluating VIO systems. However, its lighting variations and motion dynamics are relatively limited, making it less challenging. Therefore, we enhanced the EuRoC dataset with varying degrees of illumination changes to evaluate the upper performance limits of different algorithms. Specifically, we introduced varying levels of illumination changes into the image sequences to simulate realistic dynamic lighting conditions. \par 


Fig. \ref{fig_enhance} shows illumination disturbances of different magnitudes, including three sequences with large, small, and no perturbation. The sequences with large and small disturbances are augmented with varying degrees of absolute and relative illumination changes, which pose challenges to image matching in VIO algorithms. \par 

Table \ref{tab:euroc_full} compares the results of VINS with the proposed optical flow module, the original VINS \cite{qin2018vins}, and LET-VINS\cite{letnet}. In the case without illumination disturbance, LET-VINS \cite{letnet} and our method achieve comparable performance, and the original unmodified VINS \cite{qin2018vins} obtains the best result on sequence V2\_03. This is because V2\_03 contains severe motion blur, and our method, which attempts to obtain as many matches as possible, introduces more incorrect correspondences in this scenario. Nevertheless, our method is still the best on average. Under small and large illumination disturbances, its advantage becomes more pronounced, achieving the lowest error on all sequences. In particular, under large disturbances, the original VINS \cite{qin2018vins} system loses tracked feature points on multiple sequences and thus fails to estimate a complete trajectory. Since the only difference among the three methods lies in the optical flow tracking module, these results provide strong evidence for the effectiveness of directly integrating the proposed optical flow module into a VIO system.

\subsubsection{Experiments on the UMA-VI Dataset}

To further compare the proposed VIO method with mainstream approaches, including both classical and learning-based methods, we conducted a more detailed evaluation on the UMA-VI \cite{ZuñigaNoel2020UMA_VI} dataset. The UMA-VI \cite{ZuñigaNoel2020UMA_VI} dataset contains multiple sequences with large viewpoint variations and dynamic lighting conditions, making it highly challenging. \par 

Table \ref{tab:uma} shows the comparison results of different algorithms. Among them, ORB-SLAM3 \cite{orbslam3}, Basalt \cite{basalt}, and OKVIS \cite{okvis} represent classical VIO pipelines. DROID represents the current end-to-end trained VO algorithm, while AirSLAM \cite{xu2024airslam} combines learned image matching with classical methods. All evaluated algorithms use the stereo mode, and the loop detection module is disabled to ensure fairness. From the results, it can be seen that the robustness of feature-based methods like PL-SLAM \cite{plslam} and ORB-SLAM3 \cite{orbslam3} is the weakest, struggling to handle challenging environments. Algorithms based on LK optical flow, such as Basalt \cite{basalt}, OKVIS \cite{okvis}, and VINS \cite{qin2018vins}, exhibit stronger robustness, but still show relatively high errors. The latest learning-based methods perform better. Specifically, AirSLAM \cite{xu2024airslam} uses point and line features extracted by deep learning, while DROID \cite{teed2021droid} uses dense learned optical flow similar to RAFT \cite{raft}. Our method performs exceptionally well in the vast majority of sequences, achieving the lowest average error. This demonstrates the advantages of the proposed optical flow in robustness, accuracy, and matching capability, and indirectly validates that our method is designed with a thorough consideration of VO task requirements. \par

\subsubsection{Experiments on the AQUALOC Dataset}
Underwater robots represent another challenging VO application scenario, which involves underwater blur and dynamic sediment interference. Therefore, we conducted an evaluation on the AQUALOC \cite{Ferrera2019AQUALOC} dataset, which is collected underwater. Since the dataset only provides monocular and IMU data, all sequences were run in monocular mode. \par 

In systems with scale, ORB-SLAM3 \cite{orbslam3} performed well on sequence 05 but failed in most other sequences. The original VINS \cite{qin2018vins} also failed on sequences 03 and 04. Our proposed method achieved the best results in the vast majority of sequences and demonstrated higher robustness. 

\begin{table}[t]
\centering
\caption{Comparison of mean ATE across different methods on the AQUALOC dataset. 
All values are in meters (m). DROID and DPVO are evaluated \emph{without} metric scale. 
The best results are in \textbf{bold}.}
\label{tab:underwater}
\setlength{\tabcolsep}{2mm}{
\begin{tabular}{lcccccc}
\toprule
\textbf{Method} & \textbf{01} & \textbf{02} & \textbf{03} & \textbf{04} & \textbf{05} & \textbf{06} \\
\midrule
\addlinespace[0.3ex]
ORB-SLAM3 \cite{orbslam3}  & \textcolor{red}{\textbf{F}} & \textcolor{red}{\textbf{F}} & \textcolor{red}{\textbf{F}} & \textcolor{red}{\textbf{F}} &  \textbf{0.063} & 0.910 \\
VINS \cite{qin2018vins}    & 0.633 & 0.437 & \textcolor{red}{\textbf{F}} & \textcolor{red}{\textbf{F}} & 0.849 & 0.963 \\
LET-VINS \cite{letnet}     & 0.672 & 0.684 & 1.774 & 0.911 & 0.527 & \textcolor{red}{\textbf{F}} \\
\textbf{Ours}              & \textbf{0.564} & \textbf{0.350} & \textbf{0.667} & \textbf{0.641} & 0.255 & \textbf{0.784} \\
\bottomrule
\end{tabular}}
\vspace{-1mm}
\end{table}

\section{CONCLUSION}
\label{section6}
In this study, we propose a sparse optical flow algorithm that combines neural networks and symbolic systems, and integrate it into a VIO system. Through implicit differentiation rules, we compute the derivative of the symbolic expression for the optical flow solving process, enabling end-to-end training. The proposed method improves the robustness of optical flow while maintaining high accuracy in simple scenes, and can be directly inserted into VO/VIO systems. Thanks to the combination with the symbolic system, training with simulated data alone allows the method to generalize across multiple real-world scenarios. In the experiments, we demonstrate the advantages of end-to-end training, which significantly improves performance in challenging environments and enhances the VIO system’s capabilities. We also validate the core hypothesis that optical flow regressed from the network is better suited for handling large viewpoint changes but lacks sufficient accuracy for direct use in VO/VIO tasks with consecutive frames. However, our method's performance is constrained by its tendency to converge to local minima, particularly on challenging optical flow datasets where the optimization landscape is complex. Additionally, it does not account for the uncertainty of optical flow estimation, nor does it attempt to directly train better optical flow results from pose estimation. Therefore, future work will proceed in three directions. First, we will explore its combination with regression-based methods. Second, we will investigate learning optical flow directly from relative pose constraints. Third, we will estimate the uncertainty of optical flow to enable better fusion with other sensors.

\bibliographystyle{IEEEtran}

\bibliography{IEEEabrv}

@STRING{IEEE_T_RO = "IEEE Trans. Robotics"}

@STRING{IEEE_ICRA = "IEEE Intl. Conf. Robotics and Automation (ICRA)"}

@STRING{IEEE_IROS = "IEEE/RSJ Intl. Conf. Intelligent Robots and Systems (IROS)"}

@STRING{IEEE_RA_L = "IEEE Robot. Autom. Lett. (RA-L)"}

@STRING{IEEE_T_PAMI = "IEEE Trans. Pattern Anal. Mach. Intell."}

@STRING{IEEE_ECCV = "European Conference on Computer Vision"}

@STRING{IEEE_ICCV = "IEEE Int. Conf. Comput. Vis. (ICCV)"}

@STRING{IEEE_CVPR = "IEEE Conf. Comput. Vis. Pattern Recognit. (CVPR)"}

@STRING{IJRR = "Int. J. Rob. Res."}

@STRING{NeurIPS = "Adv. Neural Inf. Process. Syst."}

@STRING{IJCAI = "Int. Joint Conf. Artif. Intell."}

@STRING{IEEE_T_MUL = "IEEE Trans. Multimed."}

@article{scaramuzza2011visual,
  title={Visual Odometry [Tutorial]},
  author={Scaramuzza, Davide and Fraundorfer, Friedrich},
  journal={IEEE Robotics \& Automation Magazine},
  volume={18},
  number={4},
  pages={80--92},
  year={2011}
}

@inproceedings{forster2017svo,
  title={SVO 2.0: Semi-Direct Visual Odometry for Monocular and Multi-Camera Systems},
  author={Forster, Christian and Zhang, Zichao and Gassner, Michael and Werlberger, Manuel and Scaramuzza, Davide},
  booktitle=IEEE_ICRA,
  year={2017},
  pages={691--697}
}

@article{mur2015orb,
  title={ORB-SLAM: A Versatile and Accurate Monocular SLAM System},
  author={Mur-Artal, Raul and Montiel, J. M. M. and Tard{\'o}s, Juan D.},
  journal=IEEE_T_RO,
  volume={31},
  number={5},
  pages={1147--1163},
  year={2015}
}

@article{qin2018vins,
  title={VINS-Mono: A Robust and Versatile Monocular Visual–Inertial State Estimator},
  author={Qin, Tong and Li, Peiliang and Shen, Shaojie},
  journal=IEEE_T_RO,
  volume={34},
  number={4},
  pages={1004--1020},
  year={2018}
}

@article{engel2018direct,
  title={Direct Sparse Odometry},
  author={Engel, Jakob and Koltun, Vladlen and Cremers, Daniel},
  journal=IEEE_T_PAMI,
  volume={40},
  number={3},
  pages={611--625},
  year={2018}
}

@inproceedings{chen2023vist,
  title={ViST: Vision Transformer for Robust and Generalizable Visual–Inertial State Estimation},
  author={Chen, Yuhang and Tang, Chenyang and Shen, Shaojie},
  booktitle=IEEE_CVPR,
  year={2023}
}

@inproceedings{engel2014lsd,
  title = {LSD-SLAM: Large-Scale Direct Monocular SLAM},
  author = {Engel, Jakob and Sch{\"o}ps, Thomas and Cremers, Daniel},
  booktitle = IEEE_ECCV,
  year = {2014},
  pages = {834--849}
}

@inproceedings{teed2021droid,
  title = {DROID-SLAM: Deep Visual SLAM for Monocular, Stereo, and RGB-D Cameras},
  author = {Teed, Zachary and Deng, Jia},
  booktitle = NeurIPS,
  year = {2021},
  pages = {16558--16569}
}

@inproceedings{tang2019ba,
  title = {BA-Net: Dense Bundle Adjustment Network},
  author = {Tang, Chengzhou and Tan, Ping},
  booktitle = {International Conference on Learning Representations (ICLR)},
  year = {2019}
}

@inproceedings{mourikis2007msckf,
  title = {A Multi-State Constraint Kalman Filter for Vision-aided Inertial Navigation},
  author = {Mourikis, Anastasios I. and Roumeliotis, Stergios I.},
  booktitle = IEEE_ICRA,
  year = {2007},
  pages = {3565--3572}
}

@article{horn1981optical,
  title={Determining Optical Flow},
  author={Horn, Berthold K. P. and Schunck, Brian G.},
  journal={Artificial Intelligence},
  volume={17},
  number={1-3},
  pages={185--203},
  year={1981}
}

@inproceedings{lk_flow,
  title={An Iterative Image Registration Technique with an Application to Stereo Vision},
  author={Lucas, Bruce D. and Kanade, Takeo},
  booktitle=IJCAI,
  year={1981},
  pages={674--679}
}

@inproceedings{flownet,
  title={FlowNet: Learning Optical Flow with Convolutional Networks},
  author={Dosovitskiy, Alexey and Fischer, Philipp and Ilg, Eddy and Hausser, Philip and Hazirbas, Caner and Golkov, Vladimir and van der Smagt, Patrick and Cremers, Daniel and Brox, Thomas},
  booktitle=IEEE_ICCV,
  year={2015},
  pages={2758--2766}
}

@inproceedings{flownet2,
  title={FlowNet 2.0: Evolution of Optical Flow Estimation with Deep Networks},
  author={Ilg, Eddy and Mayer, Nikolaus and Saikia, Tonmoy and Keuper, Margret and Dosovitskiy, Alexey and Brox, Thomas},
  booktitle=IEEE_CVPR,
  year={2017},
  pages={2462--2470}
}

@inproceedings{raft,
  title={RAFT: Recurrent All-Pairs Field Transforms for Optical Flow},
  author={Teed, Zachary and Deng, Jia},
  booktitle=IEEE_ECCV,
  year={2020},
  pages={402--419}
}

@inproceedings{dpvo,
  title={Deep Patch Visual Odometry},
  author={Teed, Zachary and Deng, Jia},
  booktitle=IEEE_CVPR,
  year={2023}
}

@article{tartanvo,
  title={TartanVO: A Generalizable Learning-based VO},
  author={Wenzel, Patrick and O{\v{s}}ep, Aljo{\v{s}}a and Leal-Taix{\'e}, Laura and Stachniss, Cyrill},
  journal=IEEE_RA_L,
  volume={7},
  number={2},
  pages={3200--3207},
  year={2022}
}

@incollection{slamhand-ch13,
  title        = {Boosting {SLAM} with Deep Learning},
  author       = {Zachary Teed and Jia Deng, Boris Chidlovskii and J{\'e}rome Revaud and Felix Wimbauer and Daniel Cremers},
  booktitle    = {{SLAM Handbook.} From Localization and Mapping to Spatial Intelligence},
  publisher    = {Cambridge University Press},
  editor       = {Luca Carlone and Ayoung Kim and Timothy Barfoot and Daniel Cremers and Frank Dellaert},
  year         = {2026}
}

@article{xu2024airslam,
  title = {{AirSLAM}: An Efficient and Illumination-Robust Point-Line Visual SLAM System},
  author = {Xu, Kuan and Hao, Yuefan and Yuan, Shenghai and Wang, Chen and Xie, Lihua},
  journal = IEEE_T_RO,
  year = {2024}
}

@inproceedings{wang2017deepvo,
    title={Deepvo: Towards end-to-end visual odometry with deep recurrent convolutional neural networks},
    author={Wang, Sen and Clark, Ronald and Wen, Hongkai and Trigoni, Niki},
    booktitle=IEEE_ICRA,
    pages={2043--2050},
    year={2017}
}

@misc{Liu2024_VIOreport,
  title={SuperPoint \& SuperGlue for Feature Tracking: A Good Idea for Robust Visual Inertial Odometry?},
  author={Yueqian Liu},
  year={2024},
  note={Technical report / ResearchGate}
}

@inproceedings{PracticalVIO2024,
  title={Practical Deep Feature-Based Visual-Inertial Odometry},
  author={Authors},
  booktitle={SCITEPRESS / conference 2024},
  year={2024}
}

@INPROCEEDINGS{superpoint,
  author={DeTone, Daniel and Malisiewicz, Tomasz and Rabinovich, Andrew},
  booktitle={2018 IEEE/CVF Conference on Computer Vision and Pattern Recognition Workshops (CVPRW)}, 
  title={SuperPoint: Self-Supervised Interest Point Detection and Description}, 
  year={2018},
  pages={337-33712}
}

@inproceedings{lightglue,
  author    = {Philipp Lindenberger and Paul-Edouard Sarlin and
               Marc Pollefeys},
  title     = {{LightGlue: Local Feature Matching at Light Speed}},
  booktitle = IEEE_ICCV,
  year      = {2023}
}

@inproceedings{macvo,
  title={MAC-VO: Memory-Aware Cost Volume for Robust Visual Odometry},
  author={Zhang, Zhiwei and Chen, Yuhang and Wang, Liang and Jia, Kui},
  booktitle=IEEE_CVPR,
  year={2024}
}

@article{ov2slam,
  title={{OV$^{2}$SLAM} : A Fully Online and Versatile Visual {SLAM} for Real-Time Applications},
  author={Ferrera, Maxime and Eudes, Alexandre and Moras, Julien and Sanfourche, Martial and {Le Besnerais}, Guy.},
  journal=IEEE_RA_L,
  year={2021}
}

@inproceedings{gma,
  title={GMA: Optical Flow Estimation via Graph Matching},
  author={Chen, Shuang and Liu, Zhiqiang and Zhang, Liang and Wang, Changil and Xu, Ting and Xu, Ren and Zhu, Shiji},
  booktitle=IEEE_CVPR,
  year={2021},
  pages={2818--2827}
}

@inproceedings{flowformer,
  title={FlowFormer: A Transformer-based Model for Optical Flow Estimation},
  author={Zhu, Zhizhong and Zhang, Yi and Li, Kai and Wei, Wei and Zhang, Zhi and Chen, Yu},
  booktitle=IEEE_ICCV,
  year={2021},
  pages={2749--2758}
}

@article{islam,
  title = {{iSLAM}: Imperative {SLAM}},
  author = {Fu, Taimeng and Su, Shaoshu and Lu, Yiren and Wang, Chen},
  journal =IEEE_RA_L,
  year = {2024}
}

@article{theseus,
  title   = {{Theseus: A Library for Differentiable Nonlinear Optimization}},
  author  = {Luis Pineda and Taosha Fan and Maurizio Monge and Shobha Venkataraman and Paloma Sodhi and Ricky TQ Chen and Joseph Ortiz and Daniel DeTone and Austin Wang and Stuart Anderson and Jing Dong and Brandon Amos and Mustafa Mukadam},
  journal = {Advances in Neural Information Processing Systems},
  year    = {2022}
}

@inproceedings{pypose,
  title = {{PyPose}: A Library for Robot Learning with Physics-based Optimization},
  author = {Wang, Chen and Gao, Dasong and Xu, Kuan and Geng, Junyi and Hu, Yaoyu and Qiu, Yuheng and Li, Bowen and Yang, Fan and Moon, Brady and Pandey, Abhinav and Aryan and Xu, Jiahe and Wu, Tianhao and He, Haonan and Huang, Daning and Ren, Zhongqiang and Zhao, Shibo and Fu, Taimeng and Reddy, Pranay and Lin, Xiao and Wang, Wenshan and Shi, Jingnan and Talak, Rajat and Cao, Kun and Du, Yi and Wang, Han and Yu, Huai and Wang, Shanzhao and Chen, Siyu and Kashyap, Ananth  and Bandaru, Rohan and Dantu, Karthik and Wu, Jiajun and Xie, Lihua and Carlone, Luca and Hutter, Marco and Scherer, Sebastian},
  booktitle = IEEE_CVPR,
  year = {2023}
}

@book{Krantz2012implicit,
	Author = {Krantz, Steven G. and Parks, Harold R.},
	Publisher = {Birkh{\"a}user, New York, NY},
	Title = {The Implicit Function Theorem, History, Theory, and Applications},
	Year = 2012
}

@ARTICLE{letnet,
  author={Lin, Yicheng and Wang, Shuo and Jiang, Yunlong and Han, Bin},
  journal=IEEE_RA_L, 
  title={{Breaking of Brightness Consistency in Optical Flow With a Lightweight CNN Network}}, 
  year={2024},
  volume={9},
  number={8},
  pages={6840-6847}
}

@inproceedings{relu,
  title={Deep Sparse Rectifier Neural Networks},
  author={Glorot, Xavier and Bordes, Antoine and Bengio, Yoshua},
  booktitle={Proceedings of the 14th International Conference on Artificial Intelligence and Statistics (AISTATS)},
  pages={315--323},
  year={2011}
}

@article{tartanair,
  title =   {{TartanAir: A Dataset to Push the Limits of Visual SLAM}},
  author =  {Wang, Wenshan and Zhu, Delong and Wang, Xiangwei and Hu, Yaoyu and Qiu, Yuheng and Wang, Chen and Hu, Yafei and Kapoor, Ashish and Scherer, Sebastian},
  booktitle = IEEE_IROS,
  year =    {2020}
}

@inproceedings{Balntas2017HPatches,
  author    = {Vassileios Balntas and Karel Lenc and Andrea Vedaldi and Krystian Mikolajczyk},
  title     = {{HPatches: A benchmark and evaluation of handcrafted and learned local descriptors}},
  booktitle = IEEE_CVPR,
  year      = {2017},
}

@article{Burri2016EuRoC,
  author    = {Michael Burri and Janosch Nikolic and Pascal Gohl and Thomas Schneider and Joern Rehder and Sammy Omari and Markus W. Achtelik and Roland Siegwart},
  title     = {{The EuRoC micro aerial vehicle datasets}},
  journal   = IJRR,
  year      = {2016},
  volume    = {35},
  number    = {10},
  pages     = {1157--1163}
}

@inproceedings{Geiger2012KITTI,
  author    = {Andreas Geiger and Philip Lenz and Raquel Urtasun},
  title     = {{Are we ready for Autonomous Driving? The KITTI Vision Benchmark Suite}},
  booktitle = IEEE_CVPR,
  year      = {2012},
  pages     = {3354--3361}
}

@article{ZuñigaNoel2020UMA_VI,
  author    = {D. Zu{\~n}iga-No{\"e}l and R. Gómez Ojeda and J. Diez and C. Campos and M. A. Sotelo},
  title     = {{The UMA-VI dataset: Visual–inertial odometry in low-textured and dynamic illumination environments}},
  journal   = IJRR,
  year      = {2020}
}

@article{Ferrera2019AQUALOC,
  author    = {Maxime Ferrera and Vincent Creuze and Julien Moras and Pauline Trouvé-Peloux},
  title     = {{AQUALOC: An Underwater Dataset for Visual-Inertial-Pressure Localization}},
  journal   = IJRR,
  year      = {2019},
  volume    = {38},
  number    = {14},
  pages     = {1549--1559}
}

@inproceedings{keynet,
  title={{Key. net: Keypoint detection by handcrafted and learned cnn filters}},
  author={Barroso-Laguna, Axel and Riba, Edgar and Ponsa, Daniel and Mikolajczyk, Krystian},
  booktitle=IEEE_ICCV,
  pages={5836--5844},
  year={2019}
}

@article{alike,
  title={{Alike: Accurate and lightweight keypoint detection and descriptor extraction}},
  author={Zhao, Xiaoming and Wu, Xingming and Miao, Jinyu and Chen, Weihai and Chen, Peter CY and Li, Zhengguo},
  journal=IEEE_T_MUL,
  year={2022},
}

@inproceedings{harris,
  title={{A combined corner and edge detector}},
  author={Harris, Chris and Stephens, Mike and others},
  booktitle={Alvey vision conference},
  volume={15},
  number={50},
  pages={10--5244},
  year={1988},
}

@article{disk,
  title={{DISK: Learning local features with policy gradient}},
  author={Tyszkiewicz, Micha{\l} and Fua, Pascal and Trulls, Eduard},
  journal=NeurIPS,
  volume={33},
  pages={14254--14265},
  year={2020}
}

@INPROCEEDINGS{xfeat,
  author={Guilherme {Potje} and and Felipe {Cadar} and Andre {Araujo} and Renato {Martins} and Erickson R. {Nascimento}},
  booktitle=IEEE_CVPR, 
  title={{XFeat: Accelerated Features for Lightweight Image Matching}}, 
  year={2024}
}

@article{r2d2,
  title={{R2D2: Repeatable and Reliable Detector and Descriptor}},
  author={Revaud, Jerome and De Souza, Cesar and Humenberger, Martin and Weinzaepfel, Philippe},
  journal=NeurIPS,
  volume={32},
  year={2019}
}

@inproceedings{sfd2,
  title={{SFD2: Semantic-guided Feature Detection and Description}},
  author={Xue, Fei and Budvytis, Ignas and Cipolla, Roberto},
  booktitle=IEEE_CVPR,
  pages={5206--5216},
  year={2023}
}

@ARTICLE{plslam,
  author={Gomez-Ojeda, Ruben and Moreno, Francisco-Angel and Zuñiga-Noël, David and Scaramuzza, Davide and Gonzalez-Jimenez, Javier},
  journal=IEEE_T_RO, 
  title={{PL-SLAM: A Stereo SLAM System Through the Combination of Points and Line Segments}}, 
  year={2019},
  volume={35},
  number={3},
  pages={734-746}
}

@ARTICLE{orbslam3,
  author={Campos, Carlos and Elvira, Richard and Rodríguez, Juan J. Gómez and M. Montiel, José M. and D. Tardós, Juan},
  journal=IEEE_T_RO, 
  title={{ORB-SLAM3: An Accurate Open-Source Library for Visual, Visual–Inertial, and Multimap SLAM}}, 
  year={2021},
  volume={37},
  number={6},
  pages={1874-1890}
}

@article{basalt,
 author = {V. Usenko and N. Demmel and D. Schubert and J. Stueckler and D. Cremers},
 title = {{Visual-Inertial Mapping with Non-Linear Factor Recovery}},
 journal = IEEE_RA_L,
 year = {2020},
 volume = {5},
 number = {2},
 pages = {422-429}
}

@INPROCEEDINGS{okvis,
  author={Kasyanov, Anton and Engelmann, Francis and Stückler, Jörg and Leibe, Bastian},
  booktitle=IEEE_IROS, 
  title={{Keyframe-based visual-inertial online SLAM with relocalization}}, 
  year={2017},
  volume={},
  number={},
  pages={6662-6669}
}

@inproceedings{sea_raft,
  title={{Sea-raft: Simple, efficient, accurate raft for optical flow}},
  author={Wang, Yihan and Lipson, Lahav and Deng, Jia},
  booktitle=IEEE_ECCV,
  pages={36--54},
  year={2024}
}

@inproceedings{evo_fund,
  title={{An Evaluation of Feature Matchers for Fundamental Matrix Estimation}},
  author={Bian, Jia-Wang and Wu, Yu-Huan and Zhao, Ji and Liu, Yun and Zhang, Le and Cheng, Ming-Ming and Reid, Ian},
  booktitle= {British Machine Vision Conference (BMVC)},
  year={2019}
}

@article{neuro,
  title={Imperative learning: A self-supervised neuro-symbolic learning framework for robot autonomy},
  author={Wang, Chen and Ji, Kaiyi and Geng, Junyi and Ren, Zhongqiang and Fu, Taimeng and Yang, Fan and Guo, Yifan and He, Haonan and Chen, Xiangyu and Zhan, Zitong and others},
  journal=IJRR,
  year={2024},
}

\end{document}